\documentclass[11pt]{article}
\usepackage{url,hyperref,lineno,microtype,subcaption}
\usepackage[onehalfspacing]{setspace}
\usepackage{multirow}

\usepackage[margin=1in]{geometry}
\usepackage{url}
\usepackage{setspace}
\usepackage{subcaption} 
\usepackage{graphicx}
\usepackage{authblk}
\usepackage{amsmath}

\newcommand{\etal}{et al.\@}

\begin{document}
	\onecolumn
	
	\title {Self-attention in Vision Transformers Performs Perceptual Grouping, Not Attention} 
	\author{Paria Mehrani and John K. Tsotsos}
	\affil{Department of Electrical Engineering and Computer Science, \\
		York University, Toronto, Canada}
	\affil{\{paria, tsotsos\}@eecs.yorku.ca}
	\date{}
	\maketitle
	
	\begin{abstract}
		
		Recently, a considerable number of studies in computer vision involves deep neural architectures called vision transformers. Visual processing in these models incorporates computational models that are claimed to implement attention mechanisms. Despite an increasing body of work that attempts to understand the role of attention mechanisms in vision transformers, their effect is largely unknown. Here, we asked if the attention mechanisms in vision transformers exhibit similar effects as those known in human visual attention. To answer this question, we revisited the attention formulation in these models and found that despite the name, computationally, these models perform a special class of relaxation labeling with similarity grouping effects. Additionally, whereas modern experimental findings reveal that human visual attention involves both feed-forward and feedback mechanisms, the purely feed-forward architecture of vision transformers suggests that attention in these models will not have the same effects as those known in humans. To quantify these observations, we evaluated grouping performance in a family of vision transformers. Our results suggest that self-attention modules group figures in the stimuli based on similarity in visual features such as color. Also, in a singleton detection experiment as an instance of saliency detection, we studied if these models exhibit similar effects as those of feed-forward visual salience mechanisms utilized in human visual attention. We found that generally, the transformer-based attention modules assign more salience either to distractors or the ground. Together, our study suggests that the attention mechanisms in vision transformers perform similarity grouping and not attention. 
		
\end{abstract}

\section{Introduction}
The Gestalt principles of grouping suggest rules that explain the tendency of perceiving a unified whole rather than a mosaic pattern of parts. Gestaltists consider priors such as symmetry, similarity, proximity, continuity and closure as grouping principles that contribute to the perception of a whole. These principles which rely on input factors and the configuration of parts can be viewed as biases that result in the automatic emergence of figure and ground. To Gestalt psychologists, the perceptual organization of visual input to figure and ground was an early stage of processing prior to processes such as object recognition and attention. In fact, they posit that higher-level processes operate upon the automatically emerged figure. Some proponents of emergent intelligence go as far as to undermine the effect of attention on perceptual organization. For example, Rubin, known for his face-vase illusion, presented a paper in 1926 titled "On the Non-Existence of Attention". 

Despite the traditional Gestalt view, modern experimental evidence suggests that in addition to low-level factors, higher-level contributions can affect figure-ground organization. Specifically, experimental findings suggest that attention is indeed real and among the higher-level factors that influence figure-ground assignment~\cite{poort2012role,qiu2007figure}(See~\cite{Peterson2014} for review).  
Despite these discoveries and the enormous literature on attention (see~\cite{Itti_John2005}, for example), an  interesting development in recent years has been the introduction of deep neural architectures dubbed transformers that claim to incorporate attention mechanisms in their hierarchy~\cite{vaswani_attention_2017}. Transformers, originally introduced in the language domain, were ``based solely on attention mechanisms, dispensing with recurrence and convolutions entirely''~\cite{vaswani_attention_2017}. 

Following the success of transformers in the language domain, Dosovitskiy~\etal~\cite{vit} introduced the vision transformer (ViT), a transformer model based on self-attention mechanisms that received a sequence of image patches as input tokens. Dosovitskiy~\etal~\cite{vit} reported comparable performance of ViT to convolutional neural networks (CNNs) in image classification and concluded, similar to \cite{vaswani_attention_2017}, that convolution is not necessary for vision tasks. The reported success of vision transformers prompted a myriad of studies~\cite{cvt_2021,bhojanapalli_understanding_2021,cmt,lessIsMore,bottleneck_2021,focal_attn_2021,dascoli_convit_2021,tokens--token_2021,deit_2021,mahmood_robustness_2021,wu_visual_2021,xiao_early_2021,zhou_convnets_2021,han_survey_2022,caron_emerging_2021,swin_2021,dai_coatnet_2021,park_how_2022, zhou_understanding_2022, liu_convnet_2022,beit}. In most of these studies, the superior performance of vision transformers, their robustness~\cite{bhojanapalli_understanding_2021,mahmood_robustness_2021, naseerintriguing} and more human-like image classification behavior compared to CNNs~\cite{tuli2021convolutional} were attributed to the attention mechanisms in these architectures. Several hybrid models assigned distinct roles of feature extraction and global context integration to convolution and attention mechanisms respectively, and reported improved performance over models with only convolution or attention~\cite{xiao_early_2021,wu_visual_2021,cvt_2021,cmt,bottleneck_2021,dai_coatnet_2021,dascoli_convit_2021}. Hence, these studies suggested the need for both convolution and attention in computer vision applications. 

A more recent study by Zhou~\etal~\cite{zhou2022exploringViT}, however, reported that hybrid convolution and attention models do not ``have an absolute advantage'' compared to pure convolution or attention-based neural networks when their performance in explaining neural activities of the human visual cortex from two neural datasets was studied. Similarly, Liu~\etal~\cite{liu_convnet_2022} questioned the claims on the role of attention modules in the superiority of vision transformers by proposing steps to ``modernize'' the standard ResNet~\cite{resnet} into a new convolution-based model called ConvNeXt. They demonstrated that ConvNeXt with no attention mechanisms achieved competitive performance to state-of-the-art vision transformers on a variety of vision tasks. This controversy on the necessity of the proposed mechanisms compared to convolution adds to the mystery of the self-attention modules in vision transformers. Surprisingly, and to the best of our knowledge, no previous work directly investigated whether the self-attention modules, as claimed, implement attention mechanisms with effects similar to those reported in humans. Instead, the conclusions in previous studies were grounded on the performance of vision transformers versus CNNs on certain visual tasks. As a result, a question remains outstanding: Have we finally attained a deep computational vision model that explicitly integrates visual attention into its hierarchy? 

We address this question by revisiting two processing aspects of vision transformers. First, they formulate attention according to similarity of representations between tokens, resulting in perceptual similarity grouping. Second, with their feed-forward architecture, vision transformers resemble the traditional Gestalt view of automatic \textit{emergence} of complex features. In other words, by the virtue of their architecture, processing in vision transformers is not affected by higher-level factors and therefore, these models do not perform attention as is known in human vision. 

In a set of experiments, we evaluated these observations on various vision transformer models. Specifically, to quantify Gestalt-like similarity grouping, we introduced a grouping dataset of images with multiple shapes that shared/differed in various visual feature dimensions and measured grouping of figures in these architectures. Our results on a family of vision transformers indicate that the attention modules, as expected from the formulation, group image regions based on similarity. Our second observations indicates that if vision transformers implement attention, it can only be in the form of bottom-up attention mechanisms. To test this observation, we measured the performance of vision transformers in the task of singleton detection. Specifically, a model that implements attention is expected to almost immediately detect the pop-out, an item in the input that is visually distinct from the rest of the items. Our findings suggest that vision transformers perform poorly in that regard and even in comparison to CNN-based saliency algorithms. 

To summarize, our observations and experimental results suggest that ``attention mechanisms'' is a misnomer for computations implemented in so-called self-attention modules of vision transformers. Specifically, these modules perform similarity grouping and not attention. In fact, the self-attention modules implement a special class of in-layer lateral interactions that were missing in CNNs. Lateral interactions are known as mechanisms that counteract noise and ambiguity in the input signal~\cite{zucker1978vertical}. In light of this observation, the reported properties of vision transformers such as smoothing of feature maps~\cite{park_how_2022} and robustness~\cite{mahmood_robustness_2021,naseerintriguing} can be explained. These observations lead to the conclusion that the quest for a deep computational vision model that implements attention mechanisms has not come to an end yet.

In what follows, we will employ the terms attention and self-attention interchangeably as our focus is limited to vision transformers with transformer encoder blocks. Also, each computational component in a transformer block will be referred to as a module, for example, the attention module or the multi-layer perceptron (MLP) module. Both block and layer, then, will refer to a transformer encoder block that consists of a number of modules. 

\section{Materials and Methods}
\label{sec:methods}
In this section, we first provide a brief overview of vision transformers followed by revisiting attention formulation and the role of architecture in visual processing in these models. Then, we explain the details of the two experiments we performed in this study.

\subsection{Vision Transformers}
\label{sec:vision_transformers}
\begin{figure}[t]
	\centering
	\includegraphics[width=0.7\linewidth]{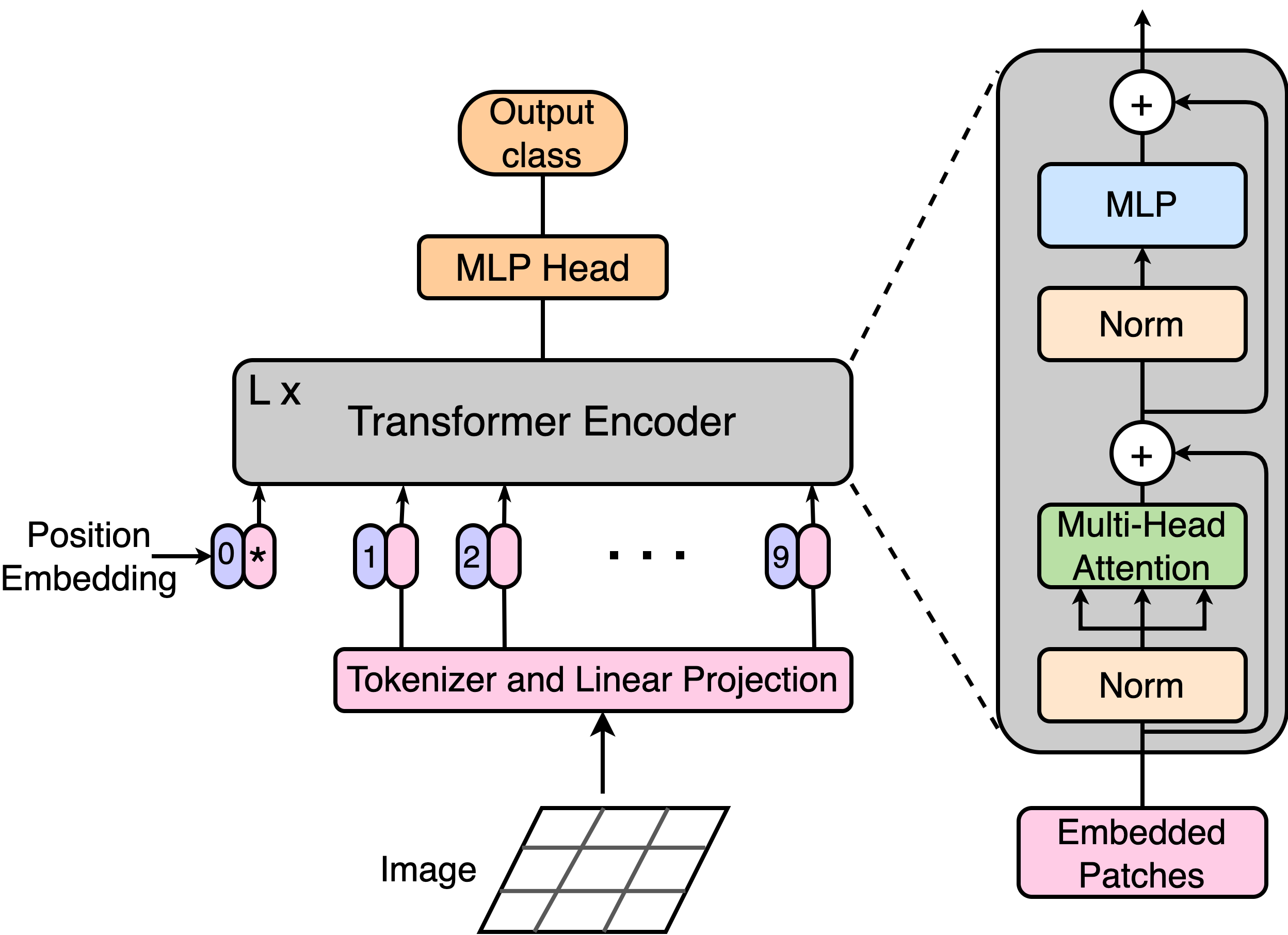}
	\caption{The ViT model architecture~\cite{vit}. First, each input image is split into local patches called tokens. After linear embedding of the tokens, a numerical position embedding is added to each token. After concatenating a learnable class embedding shown with an asterisk to the input sequence, the combined embeddings are fed to L blocks of transformer encoders. The output of the final encoder block is fed to a classification head in ViT. The zoomed-in diagram  on the right demonstrates the various modules within a transformer encoder block. These modules consist of norm, multi-head self-attention and MLP. }
	\label{fig:vit-model}
\end{figure}

Figure~\ref{fig:vit-model} provides an overview of Vision Transformer (ViT) and the various modules in its transformer encoder blocks. Most vision transformer models extend and modify or simply augment a ViT architecture into a larger system. Regardless, the overall architecture and computations in the later variants resemble those of ViT and each model consists of a number of stacked transformer encoder blocks. Each block performs visual processing of its input through self-attention, MLP and layer normalization modules. Input to these networks includes a sequence of processed image tokens (localized image patches) concatenated with a learnable class token. 

Vision transformer variants can be grouped into three main categories: 
\begin{enumerate}
	\item Models that utilized stacks of transformer encoder blocks as introduced in ViT but modified the training regime and reported a boost in performance, such as DeiT~\cite{deit_2021} and BEiT~\cite{beit}.
	\item Models that modified ViT for better adaptation to the visual domain. For example, \cite{swin_2021} introduced an architecture called Swin and suggested incorporating various scales and shifted local windows between blocks. A few other work suggested changes to the scope of attention, for example, local versus global~\cite{focal_attn_2021,chen2021crossvit}.
	\item Hybrid models that introduced convolution either as a preprocessing stage~\cite{xiao_early_2021} or as a computational step within transformer blocks~\cite{cvt_2021}.
\end{enumerate}

The family of vision transformers that we studied in our experiments includes ViT, DEiT, BEiT, Swin, and CvT. These models span all three categories of vision transformers as classified above. For each model, we studied a number of pre-trained architectures available on HuggingFace~\cite{huggingface}. Details of these architectures are outlined in Table~\ref{table:model_details}.

\begin{table}[t]
	\begin{center}
		\begin{tabular}{|c|c|c|c|c|c|}
			\hline
			Model & Architecture name & \# layers & \# params & Training dataset & Fine-tuned\\
			\hline \hline
			ViT & vit-base-patch16-224 & 12 & 86M & ImageNet-21k & --\\ 
			\hline
			\multirow{3}{*}{DeiT} & deit-tiny-distilled-patch16-224 &12 & 5M & ImageNet-1k & ImageNet-1k \\ \cline{2-6}
			& deit-small-distilled-patch16-224 & 12 & 22M & ImageNet-1k &ImageNet-1k \\ \cline{2-6}
			& deit-base-distilled-patch16-224 & 12 & 86M & ImageNet-1k & ImageNet-1k \\ \hline
			\multirow{3}{*}{BEiT} & beit-base-patch16-224 & 12 &86M& ImageNet-21k & ImageNet-1k\\ \cline{2-6}
			& beit-base-patch16-224-pt22k & 12 &86M& ImageNet-21k & --\\ \cline{2-6}
			& beit-base-patch16-224-pt22k-ft22k & 12 &86M& ImageNet-21k & ImageNet-21k\\ \hline
			\multirow{2}{*}{CvT} & cvt-13 & 13 & 19.98M & ImageNet-1k & -- \\ \cline{2-6}
			& cvt-21 & 21 & 31.54M & ImageNet-1k & --\\ \hline
			\multirow{3}{*}{Swin} & swin-tiny-patch4-window7-224 & 12 & 29M &ImageNet-1k & --\\ \cline{2-6}
			& swin-small-patch4-window7-224 & 12 & 50M&ImageNet-1k & --\\ \cline{2-6}
			\hline
		\end{tabular}
	\end{center}
	\caption{The family of vision transformers studied in this work. For each model, a number of architecture variations were studied. For all models, pre-trained architectures available on HuggingFace~\cite{huggingface} were utilized. Input resolution to all pre-trained models were $224\times224$. The datasets used for training and fine-tuning are specified. Whereas DeiT and BEiT models use the same general architecture as ViT, Swin introduces multiple scales and shifted windows to overcome the shortcomings of fixed size and position in tokens for visual tasks. The CvT architectures are hybrid models combining convolution and self-attention mechanisms in each transformer encoder block. \label{table:model_details}}
\end{table}

\subsubsection{Attention Formulation}
In transformers, the attention mechanism for a query and key-value pair is defined as:
\begin{equation}
	\text{Attention}(Q, K, V) = \text{softmax}(\frac{QK^T}{\sqrt{d_k}})V,
	\label{eq:attn}
\end{equation}
where $Q$, $K$, and $V$ represent matrices of queries, keys and values, and $\sqrt{d_k}$ is the dimension of individual key/query vectors. Vaswani~\etal~\cite{vaswani_attention_2017} explained the output of attention modules as ``a weighted sum of the values, where the weight assigned to each value is computed by a compatibility function of the query with the corresponding key.'' The same formulation was employed in ViT while the compatibility function formulation is slightly modified in some vision transformer variants. Nonetheless, the core of the compatibility function in all of these models is a dot-product measuring representation similarity.

In transformer encoders, the building block of vision transformers, the query, key and value have the same source and come from the output of the previous block. Hence, the attention modules in these blocks are called self-attention. In this case, the attention formulation can be explained as a process that results in consistent token representations across all spatial positions in the stimulus. Specifically, token representation and attention can be described as follows: each token representation signifies presence/absence of certain visual features, providing a visual interpretation or label at that spatial position. The attention mechanism incorporates the context from input into its process and describes the inter-token relations determined by the compatibility function. As a result, Equation~\ref{eq:attn} shifts the interpretation of a given token towards that of more compatible tokens in the input. The final outcome of this process will be groups of tokens with similar representations. Zucker~\cite{zucker1978vertical}referred to this process as ``Gestalt-like similarity grouping process.'' 

In (\cite{zucker1978vertical}), the Gestalt-like similarity grouping process is introduced as a type of relaxation labeling (RL) process. Relaxation labeling is a computational framework for updating the possibility of a set of labels (or interpretations) for an object based on the current interpretations among neighboring objects. Updates in RL are performed according to a \textit{compatibility function} between labels. In the context of vision transformers, at a given layer, each token is an object for which a feature representation (label) is provided from the output of the previous layer. A token representation is then updated (the residual operation after the attention module) according to a dot-product compatibility function defined between representations of neighboring tokens. In ViT, the entire stimulus forms the neighborhood for each token.  

Zucker~\cite{zucker1978vertical} defined two types of RL processes in low-level vision: vertical and horizontal. In horizontal processes, the compatibility function defines interaction at a single level of abstraction but over multiple spatial positions. In contrast, vertical processes involve interaction in a single spatial position but across various levels of abstraction. Although Zucker counts both types of vertical and horizontal processes contributing to Gestalt-like similarity grouping, self-attention formulation only fits the definition of horizontal relaxation labeling process and thus, implements a special class of RL. As a final note, while traditional RL relies on several iterations to achieve consistent labeling across all positions, horizontal processes in vision transformers are limited to a single iteration and therefore, a single iteration of Gestalt-like similarity grouping is performed in each transformer encoder block.

\subsubsection{Transformer Encoders are Feed-forward Models}
Even though the formulation of self-attention in vision transformers suggests Gestalt-like similarity grouping, this alone does not rule out the possibility of performing attention in these modules. We consider this possibility in this section. 

It is now established that humans employ a set of mechanisms, called visual attention, that limit visual processing to sub-regions of the input to manage the computational intractability of the vision problem~\cite{tsotsos1990analyzing,tsotsos2017complexity}. Despite the traditional Gestalt view, modern attention research findings suggest a set of bottom-up and top-down mechanisms determine the target of attention. For example, visual salience (``the distinct subjective perceptual quality which makes some items in the world stand out from their neighbors and immediately grab our attention''~\cite{Itti:2007}) is believed to be a bottom-up and stimulus-driven mechanism employed by the visual system to select a sub-region of the input for further complex processing. Purely feed-forward (also called bottom-up) processes, however, were shown to be facing an intractable problem with exponential computational complexity~\cite{tsotsos2021computational}. Additionally, experimental evidence suggests that visual salience~\cite{desimone95} as well as other low-level visual factors could be affected by feedback (also known as  top-down) and task-specific signals~\cite{Peterson2014,connor2004visual,baluch2011mechanisms,folk1992involuntary,bacon1994overriding,kim1999top,lamy2003does,yantis1999distinction}. In other words, theoretical and experimental findings portray an important role for top-down and guided visual processing. Finally, Herzog~\etal~\cite{herzog2014feedforward} showed how a visual processing strategy for human vision cannot be both hierarchical and strictly feed-forward through an argument that highlights the role of visual context. A literature going back to the 1800's extensively documents human attentional abilities~\cite{Itti_John2005,nobre2014oxford,carrasco25years,krauzlis2023attention,tsotsos2022attn_understand}.

Modern understanding of visual attention in humans provides a guideline to evaluate current computational models for visual attention. Vision transformers are among more recent developments that are claimed to implement attention mechanisms. However, it is evident that these models with their purely feed-forward architectures implement bottom-up mechanisms. Therefore, if it can be established that these models implement attention mechanisms, they can only capture the bottom-up signals that contribute to visual attention and not all aspects of visual attention known in humans. These observations call for a careful investigation of the effect of attention on visual processing in these models.

\subsection{Experiments}
\subsubsection{Experiment 1: Similarity Grouping}
\begin{figure}[t]
	\centering
	\includegraphics[width=\linewidth]{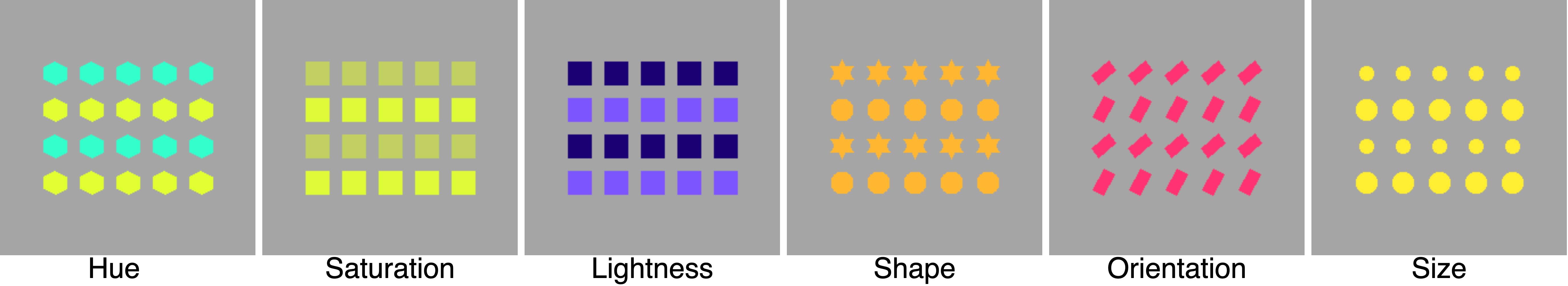}
	\caption{Similarity grouping stimuli examples. The stimuli in this dataset consists of two groups defined according to difference in one of hue, saturation, lightness, shape, orientation, size features. Each stimuli has four rows with alternating figures from the two groups. The values for the visual features that define the groups are chosen randomly. The shape set in this dataset consists of rectangle, triangle, ellipse, star, rhombus, right triangles, trapezoid, hexagon and square. Examples in this figure were picked from the version in which each figure fits within a $37\times37$ square.}
	\label{fig:grouping_dataset_examples}
\end{figure}

To quantify Gestalt-like similarity grouping in vision transformers, we created a dataset for similarity grouping with examples shown in Figure~\ref{fig:grouping_dataset_examples} and measured similarity grouping performance in vision transformers mentioned in Section~\ref{sec:vision_transformers}. As explained earlier, the attention from Equation~\ref{eq:attn} signals grouping among tokens. Therefore, we measured similarity grouping by recording and analyzing the output of attention modules in these models. 

\paragraph{Dataset}
\label{sec:grouping_dataset}
Each stimulus in the dataset consists of four rows of figures with features that differ along a single visual feature dimension including hue, orientation, lightness, shape, orientation and size. Each stimulus is $224\times224$ pixels and contains two perceptual groups of figures that alternate between the four rows. The values of the visual feature that formed the two groups in each stimulus were randomly picked. 

In some vision transformers, such as ViT, the token size and position are fixed from input and across the hierarchy. This property has been considered a shortcoming in these models when employed in visual tasks and various work attempted to address this issue~\cite{swin_2021}. Since we included vision transformers that employ ViT as their base architecture in our study, and in order to control for the token size and position in our analysis, we created the dataset such that each figure in the stimulus would fit within a single token of ViT. In this case, each figure fits a $16\times16$ pixels square positioned within ViT tokens. To measure the effect of fixed tokens on grouping, we created two other sets of stimuli. In the first set, we considered the center of every other token from ViT as a fixed position for figures and generated stimuli with figures that would fit $32\times32$ pixels squares. In this case, each figure will be relatively centered at a ViT token, but will span more than a single token. In the second set, we generated stimuli with figures that were token-agnostic. We designed these stimuli such that the set of figures was positioned at the center of the image instead of matching token positions, with each figure size fitting a $37\times37$ pixels square. 

Each version of our grouping dataset consists of 600 images with 100 stimuli per visual feature dimension, summing to a total of 1800 stimuli for all three versions. 

\paragraph{Evaluation and Metrics}
\label{sec:grouping_metrics}
For a self-attention module that yields a $H\times W\times C$ map, where $H$ and $W$ represent height and width and $C$ the number of feature channels, we first normalized the attention maps across individual feature channels so that attention scores are in the $[0, 1]$ range. Then, we measured grouping along each feature channel based on two metrics:
\begin{itemize}
	\item \textit{Grouping index:} Suppose $A_{g1}$ and $A_{g2}$ represent the average attention score of figures in group 1 and group 2 respectively. We defined the grouping index as:
	\begin{equation}
		GI = \frac{\|A_{g1} - A_{g2}\|}{A_{g1} + A_{g2}}.
	\end{equation}
	The grouping index $GI$ varies in $[0, 1]$, with larger values indicating better grouping of one group of figures in the stimulus along the feature channel.
	\item \textit{Figure-background ratio:} The overall performance of vision transformers will be impacted if background tokens are grouped with figure tokens (mixing of figure and ground). Therefore, we measured the figure-background attention ratio as:
	\begin{equation}
		AR = \max(\frac{A_{g1}}{A_{bkg}}, \frac{A_{g2}}{A_{bkg}}), 
	\end{equation}
	where $A_{g1}, A_{g2}$ represent the average attention for group 1 and group 2 figures respectively and $A_{bkg}$ is the average score of background tokens. The attention ratio $AR$ is positive and values larger than 1 indicate the attention score of at least one group of figures is larger than that of the background (the larger the ratio, the less the mixing of figure and ground). 
\end{itemize}
For each stimulus, we excluded all feature dimensions along which both $A_{g1} = 0$ and $A_{g2} = 0$ from our analysis. This happens when, for example, the feature channels represent green hues, and the figures in the stimulus are figures of red and blue. Moreover, when analyzing $AR$, we excluded all channels with $A_{bkg} = 0$ as our goal was to investigate grouping of figure and ground when some attention was assigned to the background.

\subsubsection{Experiment 2: Singleton Detection}
\label{sec:saliency_methods}
Evidence for similarity grouping in vision transformers does not indicate that self-attention modules do not perform attention. Since vision transformers are feed-forward architectures, investigating the effect of attention modules in their visual processing must be restricted to bottom-up mechanisms of attention. Therefore, we limited our study to evaluating the performance of these models in the task of singleton detection as an instance of saliency detection. Specifically, strong performance on saliency detection would suggest that these models implement the bottom-up mechanisms deployed in visual attention. 

In this experiment, we recorded the attention map of all blocks in vision transformers mentioned in Section~\ref{sec:vision_transformers}. Following \cite{BMS_saliency}, we computed an average attention map for each transformer block by averaging over all the attention channels and considered the resulting map as a saliency map. Then, we tested if the saliency map highlights the visually salient singleton. Additionally, we combined the feature maps obtained after the residual operation of attention modules and evaluated saliency detection performance for the average feature map. It is worth noting that self-attention modules, and not the features maps, are expected to highlight salient regions as the next targets for further visual processing. Nonetheless, for a better understanding of the various representations and mechanisms in vision transformers, we included feature-based saliency maps in our study. 

\begin{figure}[t!]
	\centering

\begin{subfigure}{0.9\textwidth}
		\includegraphics[width=\linewidth]{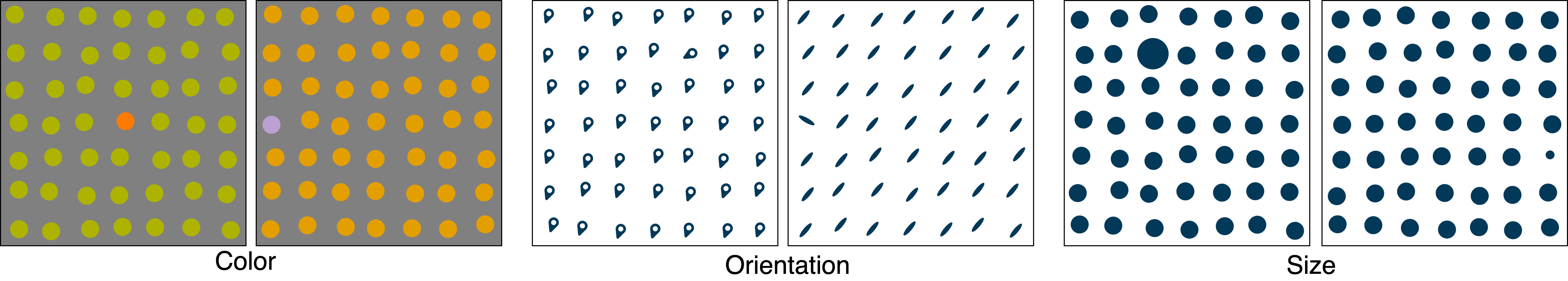}
		\caption{Examples from the Psychophysical patterns ($\text{P}^3$) dataset. The singletons in this dataset are defied according to color, orientation and size.}
		\label{subfig:P3_examples}
\end{subfigure}

\begin{subfigure}{0.9\textwidth}
		\includegraphics[width=\linewidth]{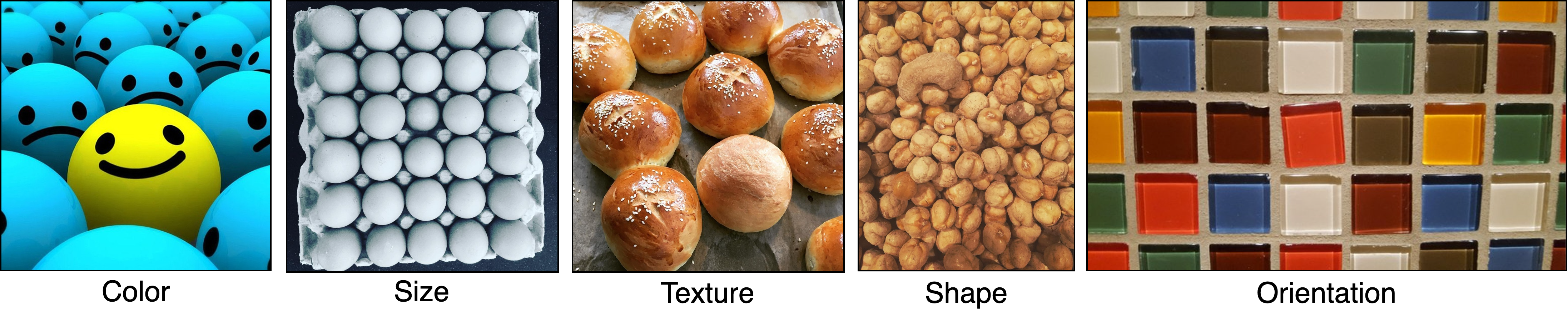}
		\caption{Examples from the Odd-one-out ($\text{O}^3$) dataset with singletons in color, size, texture, shape and orientation feature dimensions.}
		\label{subfig:O3_examples}
\end{subfigure}

	\caption{Samples of stimuli from $\text{P}^3$ and $\text{O}^3$ datasets introduced by Kotseruba~\etal~\cite{kotseruba2020saliency} are illustrated. These datasets consist of stimuli with singletons in various feature dimensions.}
	\label{fig:P3O3}
\end{figure}

\paragraph{Dataset}
For the singleton detection experiment, we utilized the psychophysical patterns ($\text{P}^3$) and odd-one-out ($\text{O}^3$) dataset introduced by Kotseruba~\etal~\cite{kotseruba2020saliency}. Examples of each set are shown in Figure~\ref{fig:P3O3}. The $\text{P}^3$ dataset consists of 2514 images of size 1024$\times$1024. Each image consists of figures on a regular $7\times7$ grid with one item as the target that is visually different in one of color, orientation or size from other items in the stimulus. The location of the target is chosen randomly. The $\text{O}^3$ dataset includes 2001 images with the largest dimension set to 1024. In contrast to the grouping and $\text{P}^3$ datasets whose stimuli were synthetic images, the $\text{O}^3$ dataset consists of natural images. Each image captures a group of objects that belong to the same category with one that stands out (target) from the rest (distractors) in one or more visual feature dimensions (color, texture, shape, size, orientation, focus and location). The $\text{O}^3$ with natural images provides the opportunity to investigate the performance of the vision transformer models in this study on the same type of stimuli those were trained. Both $\text{P}^3$ and $\text{O}^3$ datasets are publicly available and further details of both datasets can be found in~\cite{kotseruba2020saliency}.

\paragraph{Metrics}
We followed~\cite{kotseruba2020saliency} to measure singleton detection performance in vision transformers. We employed their publicly available code for the computation of metrics they used to study traditional and deep saliency models. The number of fixation and saliency ratio were measured for $\text{P}^3$ and $\text{O}^3$ images respectively as explained below.
\begin{itemize}
	\item \textit{Number of fixations:} Kotseruba~\etal~\cite{kotseruba2020saliency} used the number of fixations required to detect pop-out as a proxy for salience. Specifically, they iterated through the maxima of the saliency map until the target was detected or a maximum number of iterations was reached. At each iteration that resembles a fixation of the visual model on a region of input, they suppressed the fixated region with a circular mask before moving the fixation to the next maxima. Lower number of fixations indicates higher relative salience of the target to that of distractors.
	\item \textit{Saliency ratio:} Kotseruba~\etal~\cite{kotseruba2020saliency} employed the ratio of the maximum saliency of the target versus the maximum saliency of the distractors. They also measured the maximum saliency of the background to the maximum saliency of the target. These two ratios that are referred to as $MSR_{targ}$ and $MSR_{bg}$ determine if the target is more salient than the distractors or the background respectively. Ideally, $MSR_{targ}$ is larger than 1 and $MSR_{bg}$ is less than 1.
\end{itemize}

\section{Results}
\subsection{Experiment 1: Similarity Grouping}
Each vision transformer in our study consists of a stack of transformer encoder blocks. In this experiment, our goal was to investigate similarity grouping in attention modules in transformer encoder blocks. We were also interested in changes in similarity grouping over the hierarchy of transformer encoders. Therefore, for each vision transformer, we took the following steps: We first isolated transformer encoders in the model and computed the grouping index ($GI$) and attention ratio ($AR$) per channel as explained in Section ~\ref{sec:grouping_metrics}. Then, we considered the mean $GI$ and $AR$ per block as the representative index and ratio of the layer.

\begin{figure}
	\centering
\begin{subfigure}{0.48\textwidth}
		\includegraphics[width=\linewidth]{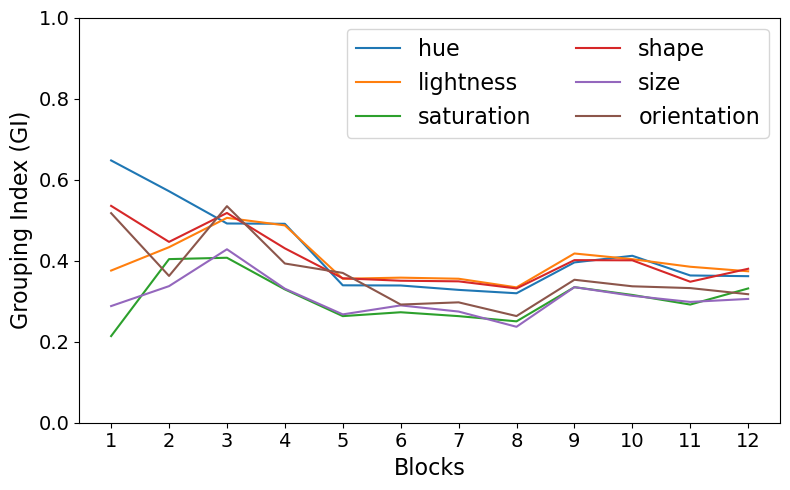}
		\caption{Mean grouping index ($GI$).}
		\label{subfig:vit-GI}
	\end{subfigure}
	\begin{subfigure}{0.48\textwidth}
		\includegraphics[width=\linewidth]{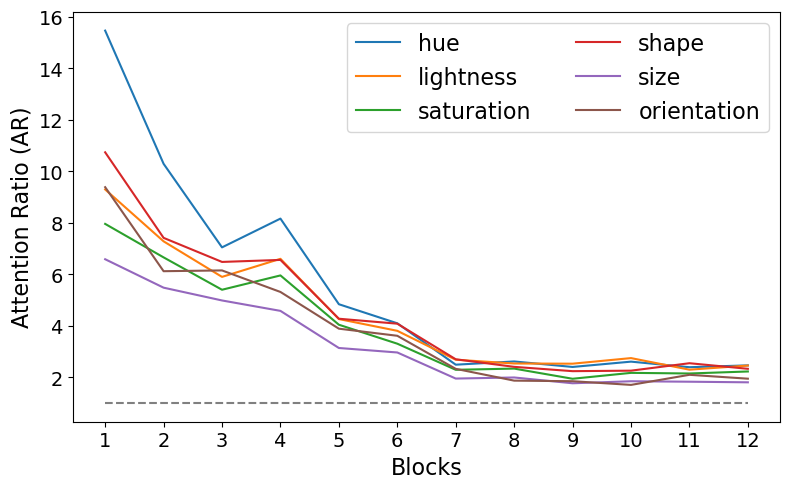}
		\caption{Mean attention ratio ($AR$).}
		\label{subfig:vit-AR}
	\end{subfigure}
	\caption{Mean grouping index and attention ratio for the vit-base-patch16-224 architecture over all stimuli but separated according to the visual features that defined the groups of figures in the input. \textbf{(A)} The mean grouping index is larger than 0.2 for all layers of the model across all visual features, suggesting perceptual grouping based on similarity in this architecture. \textbf{(B)} The attention ratio of larger than 1 for all transformer encoder blocks of vit-base-patch16-224 indicates larger scores are assigned to figure tokens. However, the steep decline in the $AR$ ratio in the hierarchy demonstrates mixing of figure and background tokens due to similar attention scores.}
	\label{fig:vit-GI-AR}
\end{figure}

Figure\ref{subfig:vit-GI} shows the mean $GI$ for the architecture called ``vit-base-patch16-224'' in Table~\ref{table:model_details} over all layers of the hierarchy. The $GI$ is plotted separately according to the visual feature that differed between the groups of figures. This plot demonstrates that $GI$ for all blocks of this model across all tested feature dimensions is distinctly larger than 0, suggesting similarity grouping of figures in all attention modules of this architecture. Interestingly, despite some variations in the first block, all layers have relatively similar $GI$. Moreover, the grouping indices for all feature dimensions are close, except for hue with $GI$ larger than 0.6 in the first block, indicating stronger grouping among tokens based on this visual feature. 

Figure~\ref{subfig:vit-AR} depicts the mean $AR$ for the same architecture, vit-base-patch16-224, for all the encoder blocks. Note that all curves in this plot are above the $AR=1$ line denoted as a dashed gray line, indicating that all attention modules assign larger attention scores to at least one group of figures in the input versus the background tokens. However, notable is the steep decline in the mean $AR$ across the hierarchy. This observation confirms the previous reports of smoother attention maps in higher stages of the hierarchy~\cite{park_how_2022} with similar attention assigned to figure and background tokens.

\begin{figure}
	\centering
	\begin{subfigure}{0.98\textwidth}
		\includegraphics[width=\linewidth]{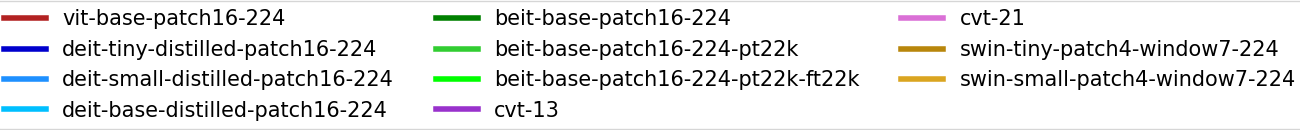}
	\end{subfigure}  
	
	\begin{subfigure}{0.49\textwidth}
		\includegraphics[width=\linewidth]{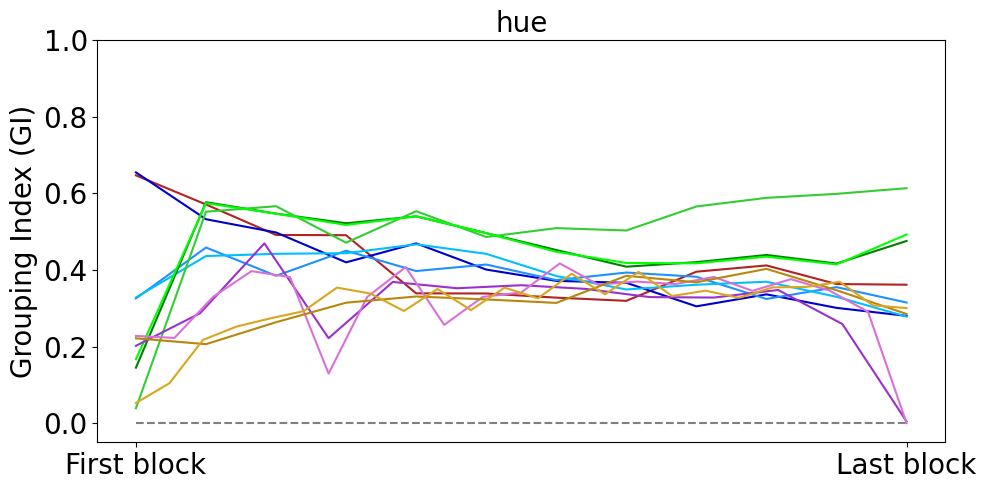}
	\end{subfigure}
	\begin{subfigure}{0.49\textwidth}
		\includegraphics[width=\linewidth]{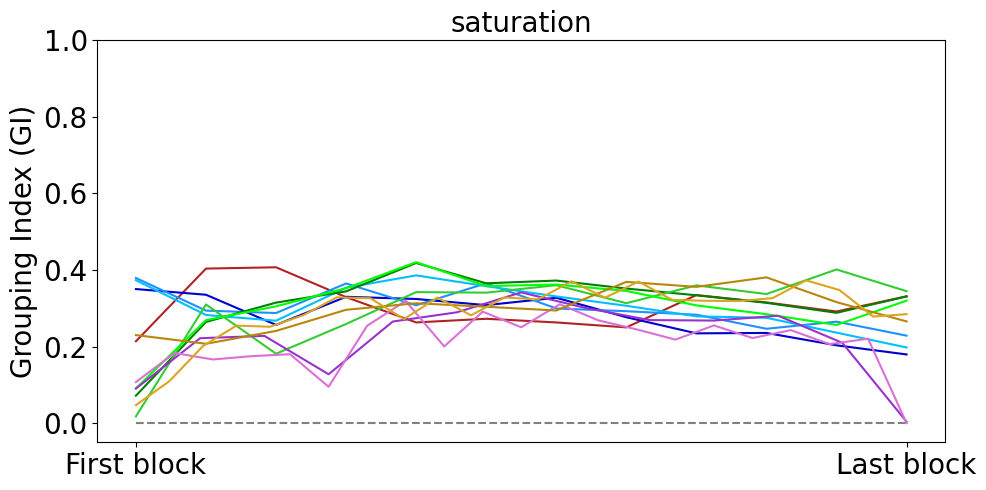}
	\end{subfigure}
	
	\begin{subfigure}{0.49\textwidth}
		\includegraphics[width=\linewidth]{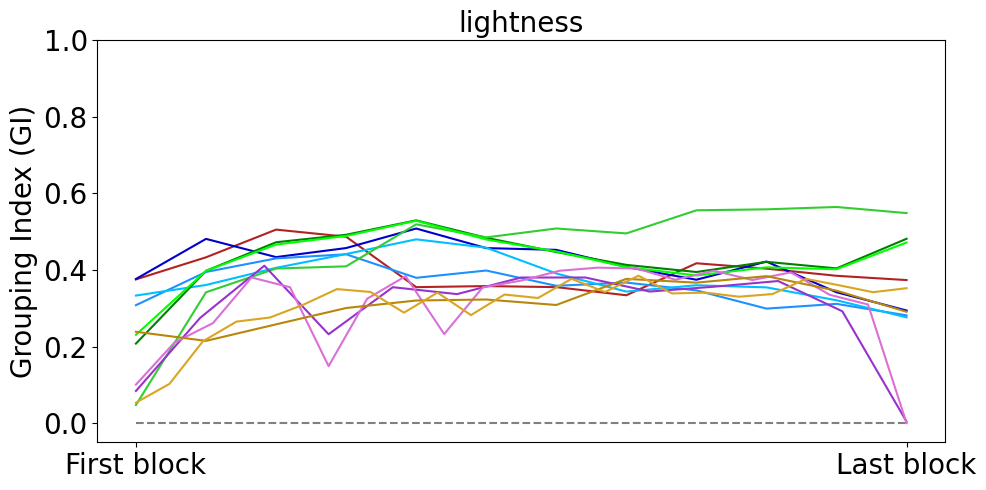}
	\end{subfigure}
	\begin{subfigure}{0.49\textwidth}
		\includegraphics[width=\linewidth]{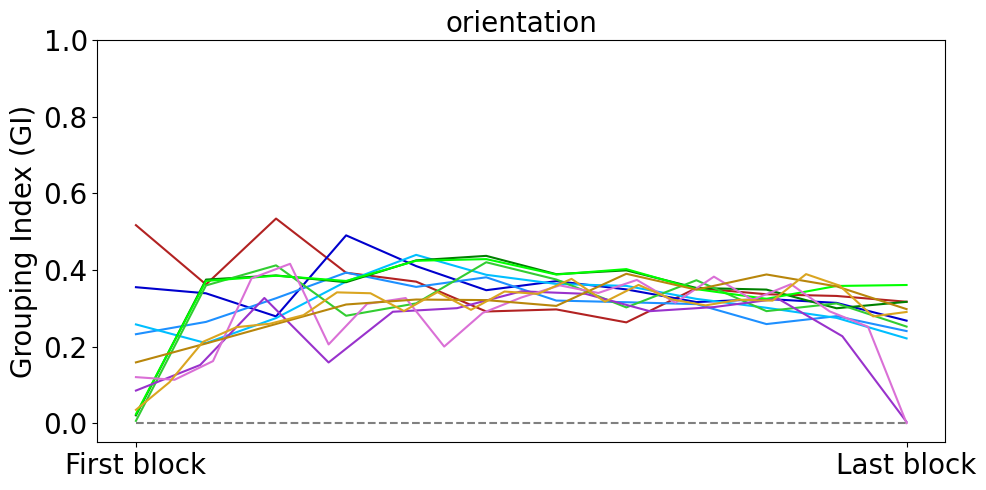}
	\end{subfigure}
	
	\begin{subfigure}{0.49\textwidth}
		\includegraphics[width=\linewidth]{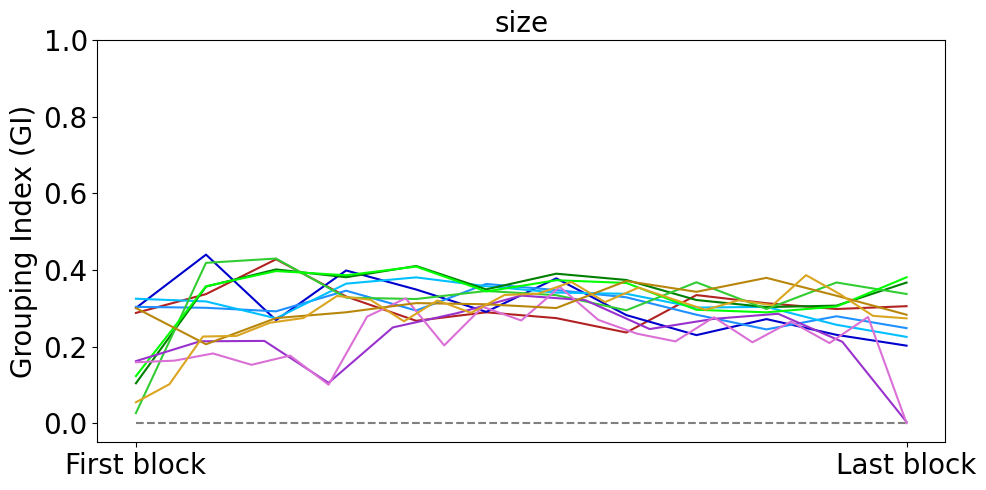}
	\end{subfigure}
	\begin{subfigure}{0.49\textwidth}
		\includegraphics[width=\linewidth]{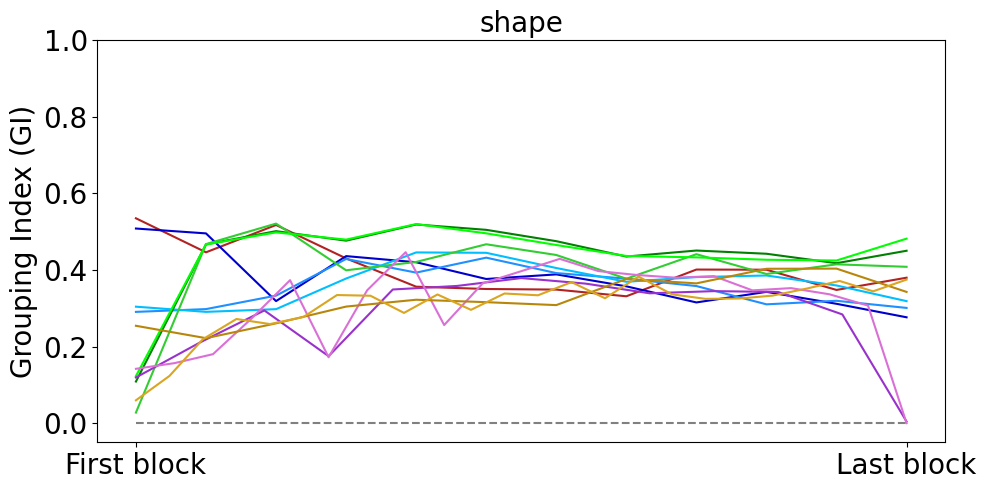}
	\end{subfigure}
	\caption{Mean grouping index for all the architectures from Table~\ref{table:model_details} plotted separately for each visual feature that defined the perceptual grouping of figures in the input. For comparison purposes, we plotted the grouping index of all models within the same range on the x-axis. Accordingly, we labeled the \textit{first} and \textit{last} block on the horizontal axis. These plots demonstrate that all models perform similarity grouping of figures based on low-level visual features such as hue and orientation. Except for the final block of cvt models, all layers of all architectures have mean $GI$ higher than 0. The legend at the top applies to all plots.}
	\label{fig: all_models_GI}
\end{figure}

Figure~\ref{fig: all_models_GI} shows the mean $GI$ for all the architectures from Table~\ref{table:model_details} separately based on the visual feature that defined the groups in the input. All models, across all their layers, with some exceptions, demonstrate mean $GI$ that are distinctly larger than 0. The exceptions include the first layer of all beit architectures and swin-small-patch4-window7-224, and the last block of cvt-13 and cvt-14. Interestingly, beit and swin architectures jump in their mean $GI$ in their second block. Even though deit and beit architectures utilized the same architecture as ViT but trained the model with more modern training regimes, both models demonstrate modest improvement over vit-base-patch16-224.

\begin{figure}
	\centering
	\begin{subfigure}{0.9\textwidth}
		\includegraphics[width=\linewidth]{legend.png}
	\end{subfigure}
	
	\begin{subfigure}{0.49\textwidth}
		\includegraphics[width=\linewidth]{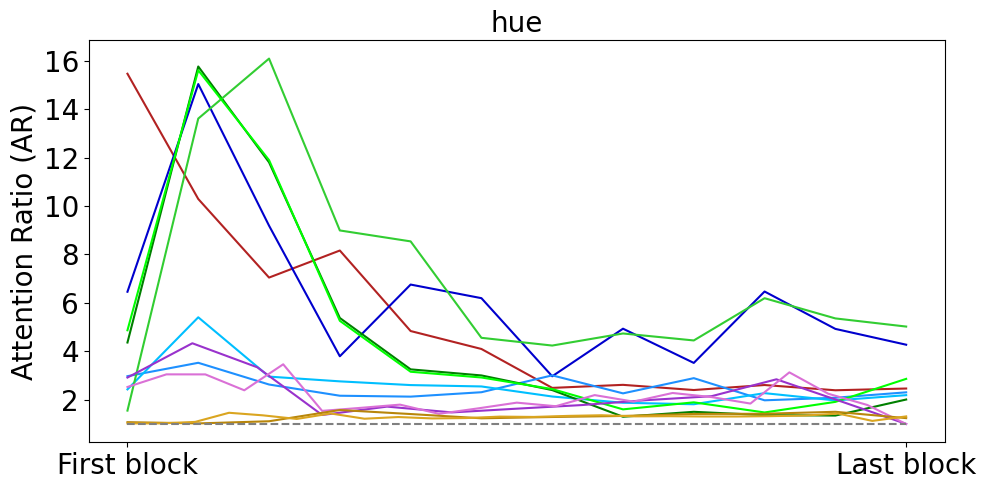}
	\end{subfigure}
	\begin{subfigure}{0.49\textwidth}
		\includegraphics[width=\linewidth]{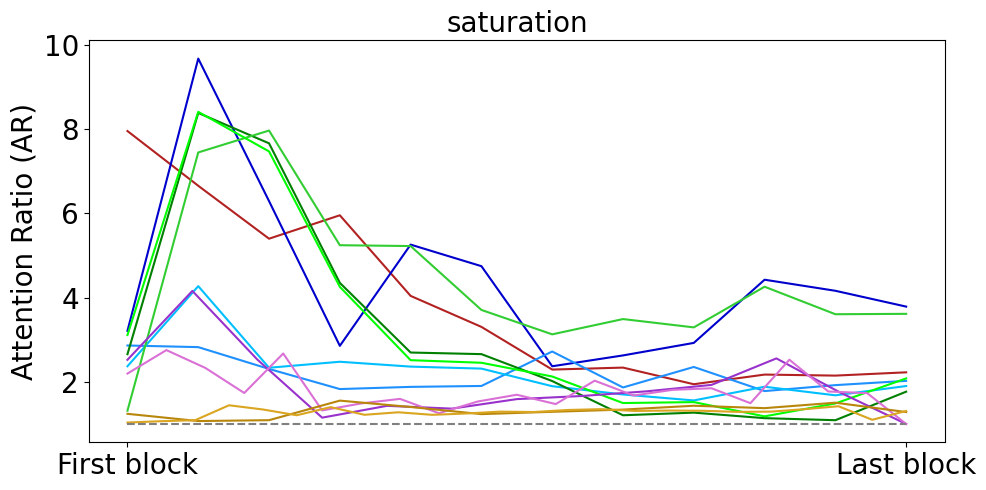}
	\end{subfigure}
	
	\begin{subfigure}{0.49\textwidth}
		\includegraphics[width=\linewidth]{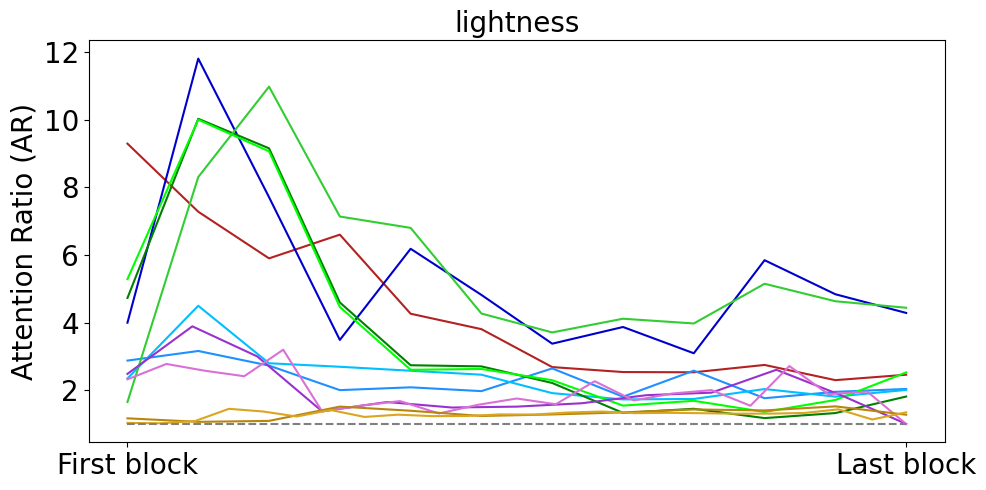}
	\end{subfigure}
	\begin{subfigure}{0.49\textwidth}
		\includegraphics[width=\linewidth]{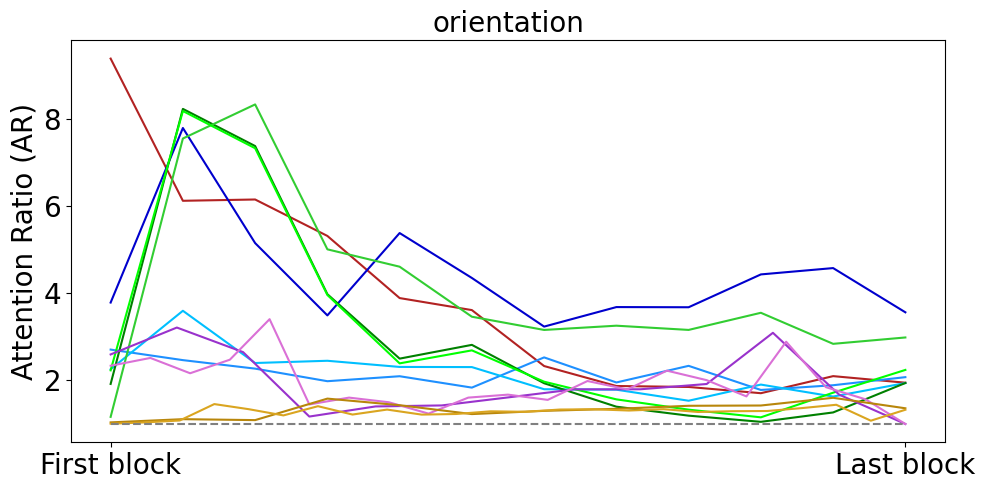}
	\end{subfigure}
	
	\begin{subfigure}{0.49\textwidth}
		\includegraphics[width=\linewidth]{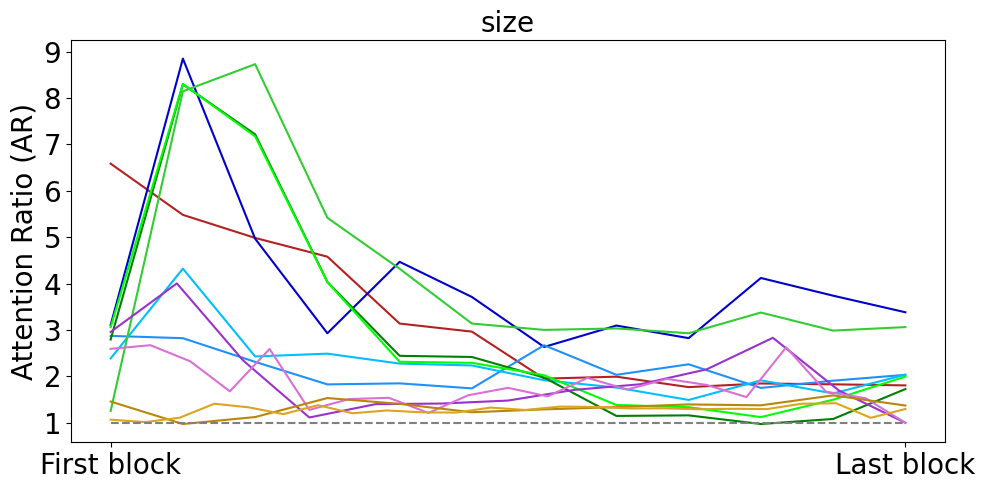}
	\end{subfigure}
	\begin{subfigure}{0.49\textwidth}
		\includegraphics[width=\linewidth]{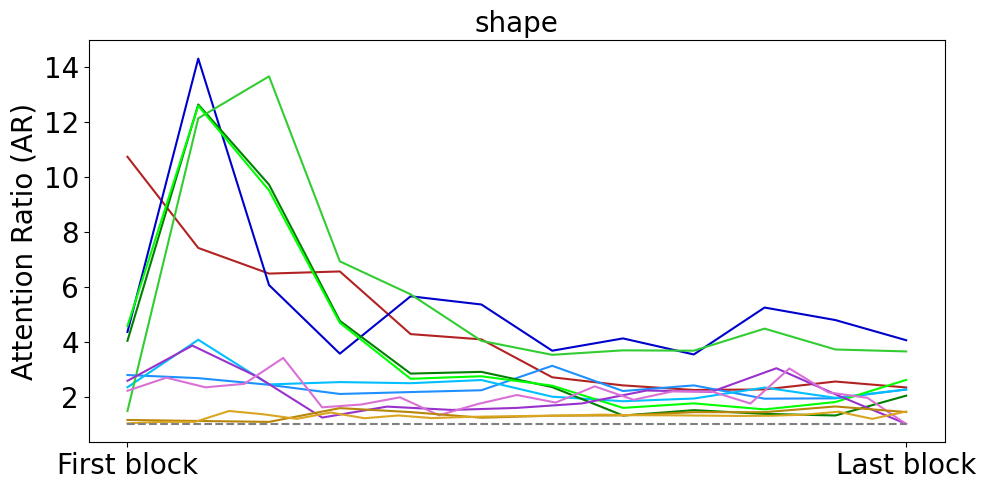}
	\end{subfigure}
	\caption{Mean attention ratio for all the architectures from Table~\ref{table:model_details} plotted separately for each visual feature that defined the perceptual grouping of figures in the input. Similar to Figure~\ref{fig: all_models_GI}, and for ease of comparison, we plotted the $AR$ for all models within the same range on the x-axis. Interestingly, swin and cvt, two models that adapted ViT to the visual domain, have relatively smaller attention ratios compared to the rest of the architectures, suggesting that incorporating scale and shifting the token position in Swin and convolution in CvT architectures results in mixing of figure and background representations and consequently attention scores. Among the other architectures that use ViT as the underlying model, the attention ratio plots are somewhat similar to those of Figure~\ref{subfig:vit-AR}, that is, larger attention ratios in earlier blocks with a decline in the hierarchy. }
	\label{fig: all_models_AR}
\end{figure}

Plots in Figure~\ref{fig: all_models_AR} depict the mean $AR$ over all the architectures. Interestingly, vit-base-patch16-224 is the only architecture whose mean $AR$ for the first block is the largest in its hierarchy and unanimously for all visual features. Among the three DeiT architectures (tiny, small, and base), deit-tiny-distilled-patch16-224, demonstrates larger mean AR ratios. Compared to ViT, deit-tiny-distilled-patch16-224 has far less parameters and the comparable mean AR for this architecture with Vit confirms the suggestion of \cite{deit_2021} that an efficient training regime in a smaller model could result in performance gain against a larger model. Results from Figure~\ref{fig: all_models_AR} are also interesting in that all of swin and cvt architectures that are claimed to adapt transformer models to the vision domain, have relatively small mean $AR$ over their hierarchy. These results show that these models mix figure and background tokens in their attention score assignments, an observation that deserves further investigation in a future work.

\begin{figure}
	\centering
	\begin{subfigure}{0.49\textwidth}
		\includegraphics[width=\linewidth]{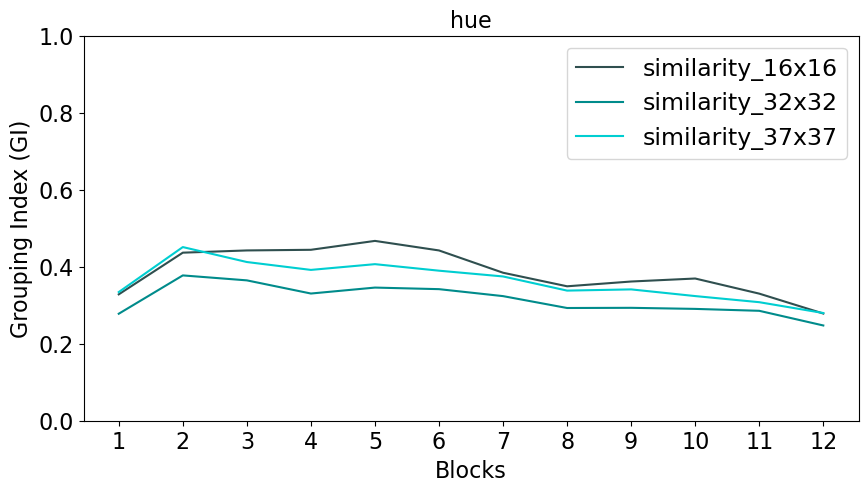}
	\end{subfigure}
	\begin{subfigure}{0.49\textwidth}
		\includegraphics[width=\linewidth]{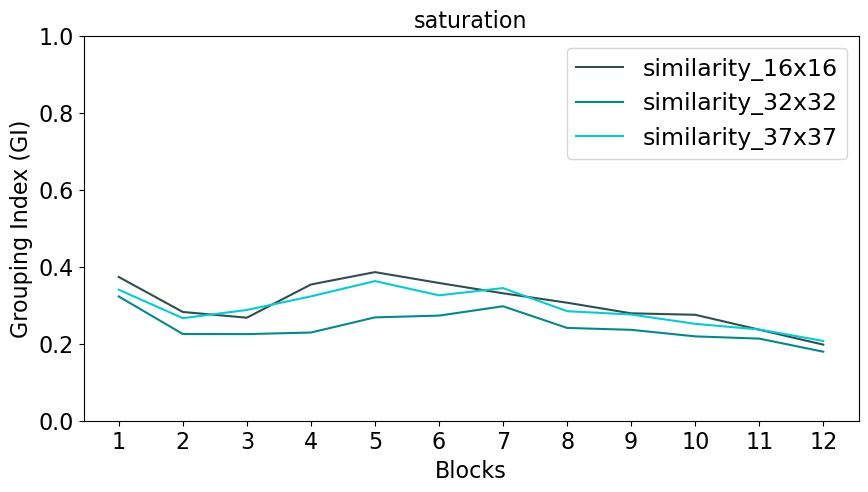}
	\end{subfigure}
	
	\begin{subfigure}{0.49\textwidth}
		\includegraphics[width=\linewidth]{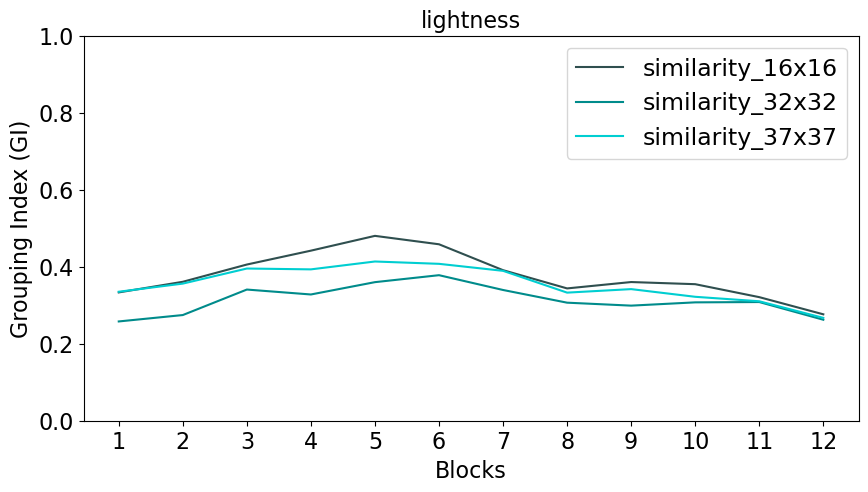}
	\end{subfigure}
	\begin{subfigure}{0.49\textwidth}
		\includegraphics[width=\linewidth]{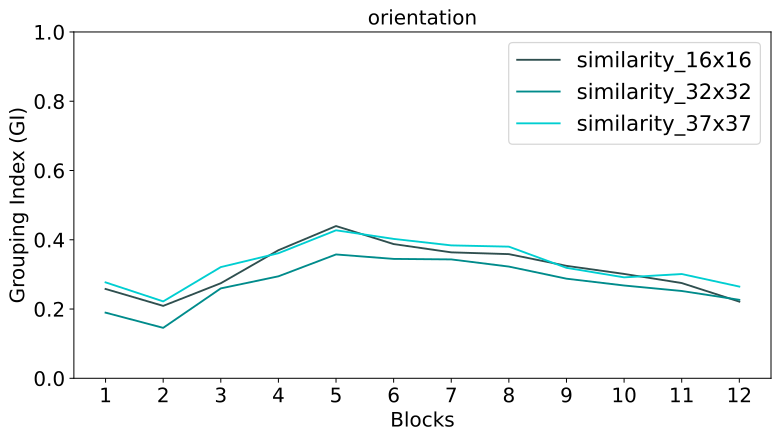}
	\end{subfigure}
	
	\begin{subfigure}{0.49\textwidth}
		\includegraphics[width=\linewidth]{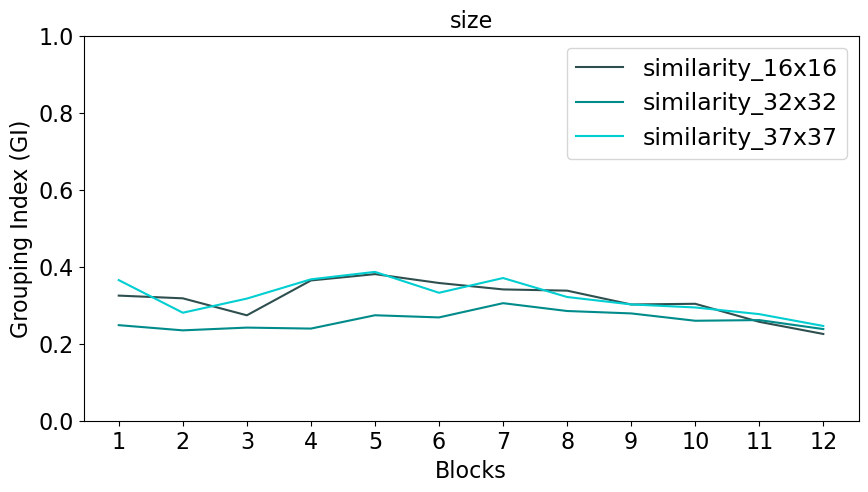}
	\end{subfigure}
	\begin{subfigure}{0.49\textwidth}
		\includegraphics[width=\linewidth]{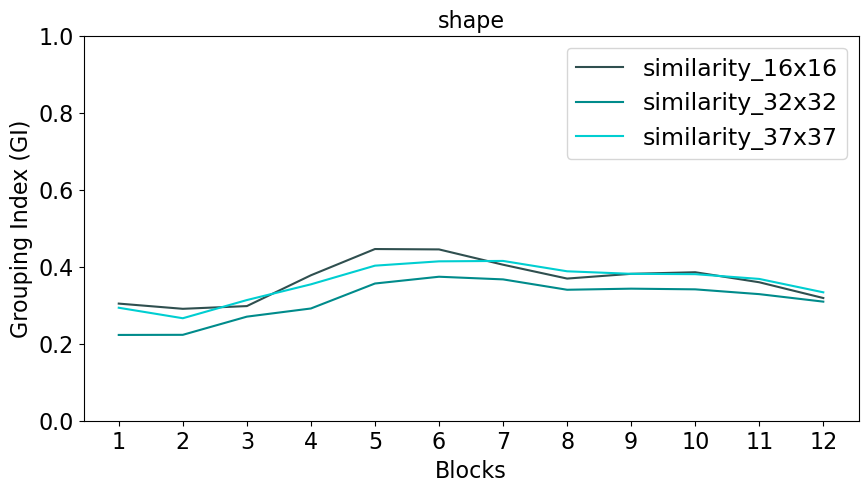}
	\end{subfigure}
	\caption{Each token in ViT-based architecturs has a fixed position and size across the hierarchy of transformer encoders. This property is noted as a shortcoming of some vision transformers. To control for position and size of tokens in these models, we designed our grouping dataset according to the ViT model tokens such that each figure in the stimulus would fit within and positioned inside the model $16\times16$ tokens. To test for the effect of figure size, we introduced a version of grouping dataset with figures centered at every other ViT token but larger in size such that each figure would fit a $32\times32$ square. We also introduced a third version where figures in the stimuli were token-agnostic. In the third version, the set of figures occupy the center of image and each figure fits within a $37\times 37$ square. We tested the grouping performance of the deit-base-distilled-patch16-224 architecture over all three versions of the dataset. Note that deit-base-distilled-patch16-224 utilizes an identical architecture as vit-base-patch16-224 with a different training regime. Our results over the various visual features in the dataset demonstrate comparable results over the three versions of the dataset, suggesting no significant effect of token position or size in grouping in vision transformers.}
	\label{fig: GI-grouping-versions}
\end{figure}

Finally, Figure~\ref{fig: GI-grouping-versions} summarizes the mean grouping index $GI$ for the deit-base-distilled-patch16-224 architecture over the three versions of the grouping dataset as explained in Section~\ref{sec:grouping_dataset}. These results demonstrate similar grouping index over all three versions, suggesting little impact of token position and size relative to figures in the input.

\subsection{Experiment 2: Singleton Detection}
Generally, in saliency experiments, the output of the model is considered for performance evaluation. In this study, however, not only we were interested in the overall performance of vision transformers (the output of the last block), but also in the transformation of the saliency signal in the hierarchy of these models. Examining the saliency signal over the hierarchy of transformer blocks would provide valuable insights into the role of attention modules in saliency detection. Therefore, we measured saliency detection in all transformer blocks. 

\subsubsection{The $\text{P}^3$ Dataset Results}
Following Kotseruba~\etal~\cite{kotseruba2020saliency}, to evaluate the performance of vision transformer models on the $\text{P}^3$ dataset, we measured the target detection rate at 15, 25, 50, and 100 fixations. Chance level performance for vit-base-patch16-224, as an example, would be 6\%, 10\%, 20\% and 40\% for 15, 25, 50 and 100 fixations respectively (masking after each fixation explained in Section~\ref{sec:saliency_methods} masks an entire token). Although these levels for the various models would differ due to differences in token sizes and incorporating multiple scales, these chance level  performances from  vit-base-patch16-224 give a baseline for comparison.

\begin{figure}
	\centering
	\begin{subfigure}{0.75\textwidth}
		\includegraphics[width=\linewidth]{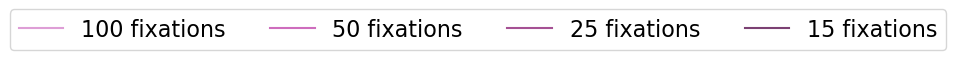}
	\end{subfigure}
	
	\begin{subfigure}{0.42\textwidth}
		\includegraphics[width=\linewidth]{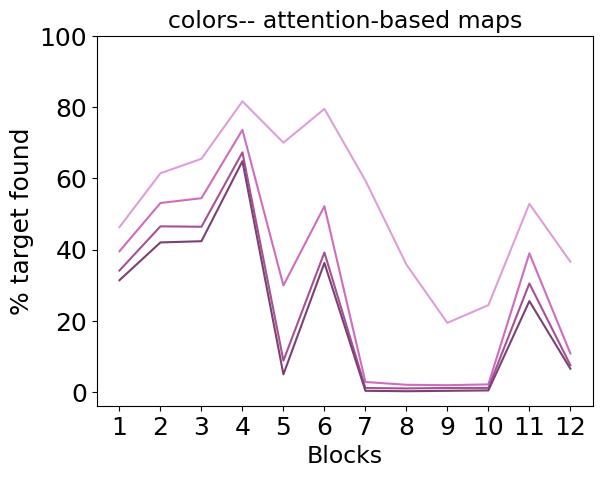}
	\end{subfigure}
	\begin{subfigure}{0.42\textwidth}
		\includegraphics[width=\linewidth]{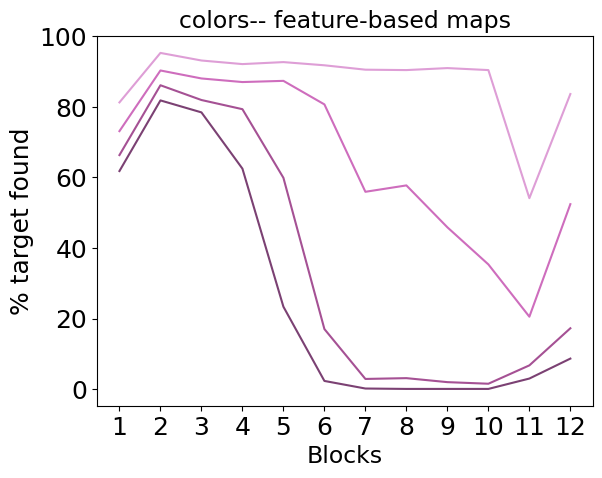}
	\end{subfigure}
	
	\begin{subfigure}{0.42\textwidth}
		\includegraphics[width=\linewidth]{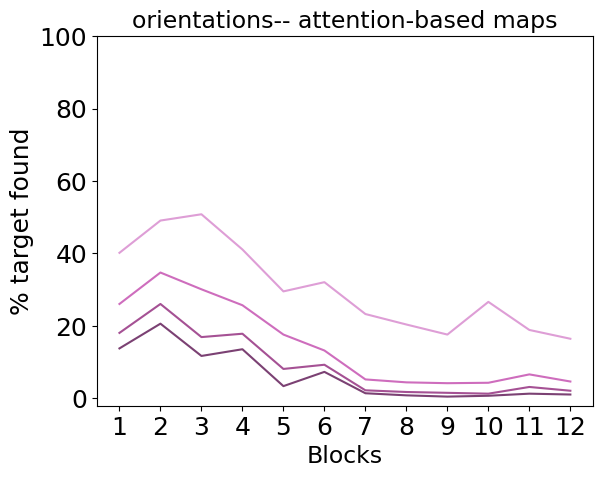}
	\end{subfigure}
	\begin{subfigure}{0.42\textwidth}
		\includegraphics[width=\linewidth]{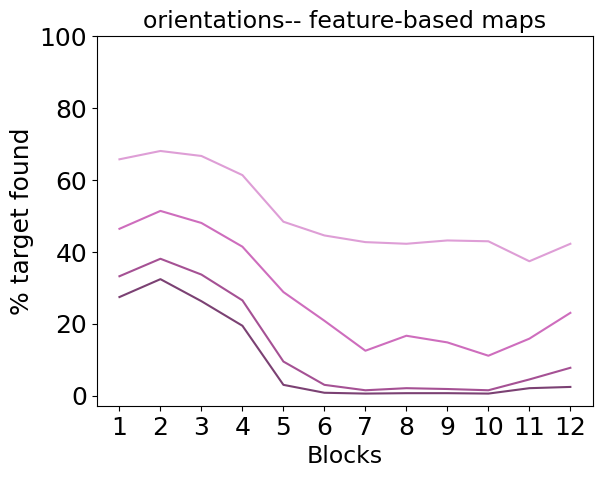}
	\end{subfigure}
	
	\begin{subfigure}{0.42\textwidth}
		\includegraphics[width=\linewidth]{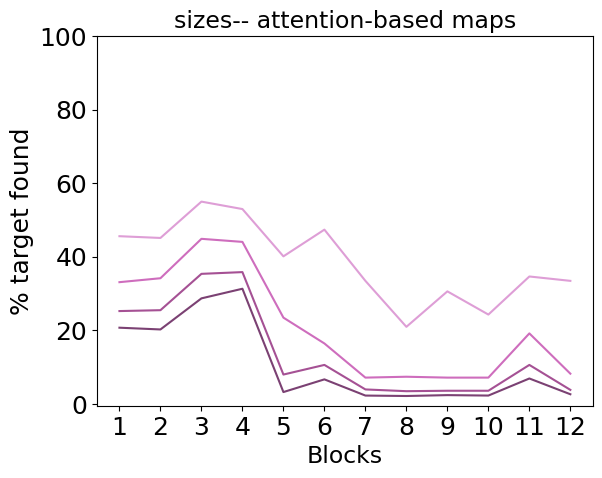}
	\end{subfigure}
	\begin{subfigure}{0.42\textwidth}
		\includegraphics[width=\linewidth]{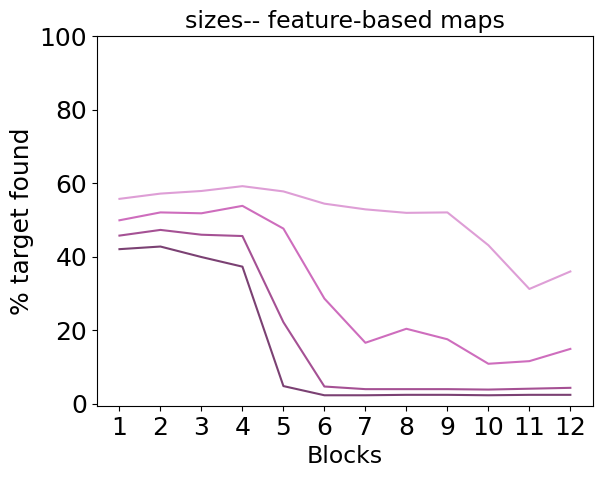}
	\end{subfigure}
	\caption{Target detection rate of the vit-base-patch16-224 model for 15, 25, 50, and 100 fixations on images of the $\text{P}^3$ dataset. Legend on top applies to all plots. For this model with $16\times16$ pixels tokens, each masking after a fixation masks almost an entire token. Therefore, chance performance will be at 6\%, 10\%, 20\% and 40\% for 15, 25, 50 and 100 fixations. Comparing the plots of the left column for attention-based saliency maps versus those on the right obtained from feature-based saliency maps indicates superior performance of feature-based maps for salient target detection. This is interesting in that modules claimed to implement attention mechanisms are expected to succeed in detecting visually salient figures in the input. Overall, for both attention and feature-based maps, color targets have higher detection rates versus orientation and size, the conditions in which performance is mainly at chance level for all fixation thresholds and across all blocks in the ViT hierarchy. Additionally, in both attention and feature-based maps, performance peaks in earlier blocks and declines in later layers, suggesting multiple transformer encoder blocks mix representations across spatial locations such that the model cannot detect the visually salient target almost immediately or even by chance. }
	\label{fig:P3-vit}
\end{figure}

Figure~\ref{fig:P3-vit} demonstrates the performance of saliency maps obtained from attention and feature maps of all vit-base-patch16-224 blocks. These plots clearly demonstrate that the feature-based saliency maps in each block outperform those computed from the attention maps. This is somewhat surprising since as explained in Section~\ref{sec:saliency_methods}, if vision transformers implement attention mechanisms, attention modules in these models are expected to highlight salient regions in the input for further visual processing. Nonetheless, plots in Figure~\ref{fig:P3-vit} tell a different story, namely that feature maps are preferred options for applications that require singleton detection. Comparing target detection rates across color, orientation and size for both attention and feature maps demonstrate higher rates in detecting color targets compared to size and orientation. For all three of color, orientation and size, the target detection rates peak at earlier blocks for attention-based saliency maps and decline in later blocks, with lower than chance performance for most blocks. This pattern is somewhat repeated in feature-based saliency maps with more flat curves in the hierarchy, especially for a larger number of fixations. 

\begin{figure}
	\centering
	\begin{subfigure}{0.8\textwidth}
		\includegraphics[width=\linewidth]{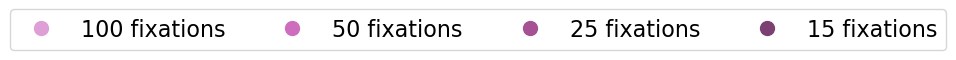}
	\end{subfigure}  
	
	\begin{subfigure}{0.49\textwidth}
		\includegraphics[width=\linewidth]{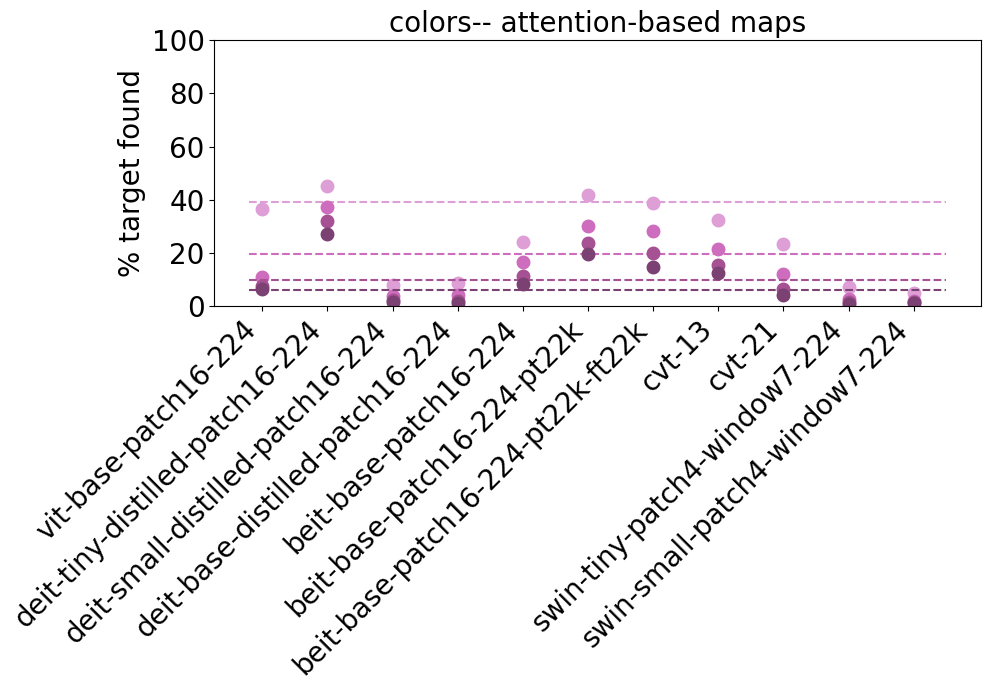}
	\end{subfigure}
	\begin{subfigure}{0.49\textwidth}
		\includegraphics[width=\linewidth]{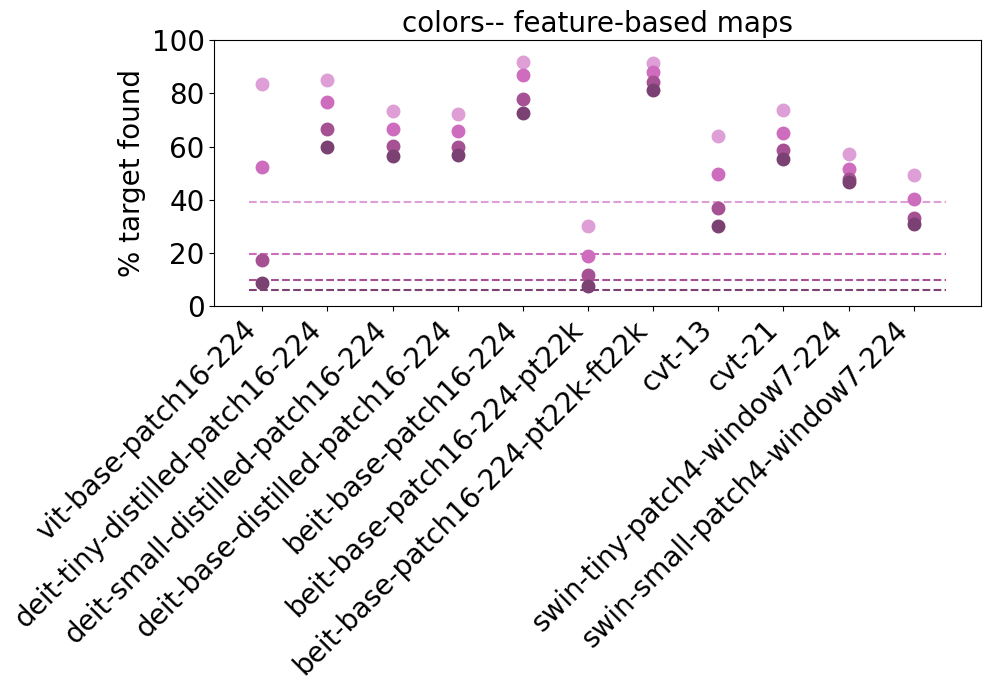}
	\end{subfigure}
	
	\begin{subfigure}{0.49\textwidth}
		\includegraphics[width=\linewidth]{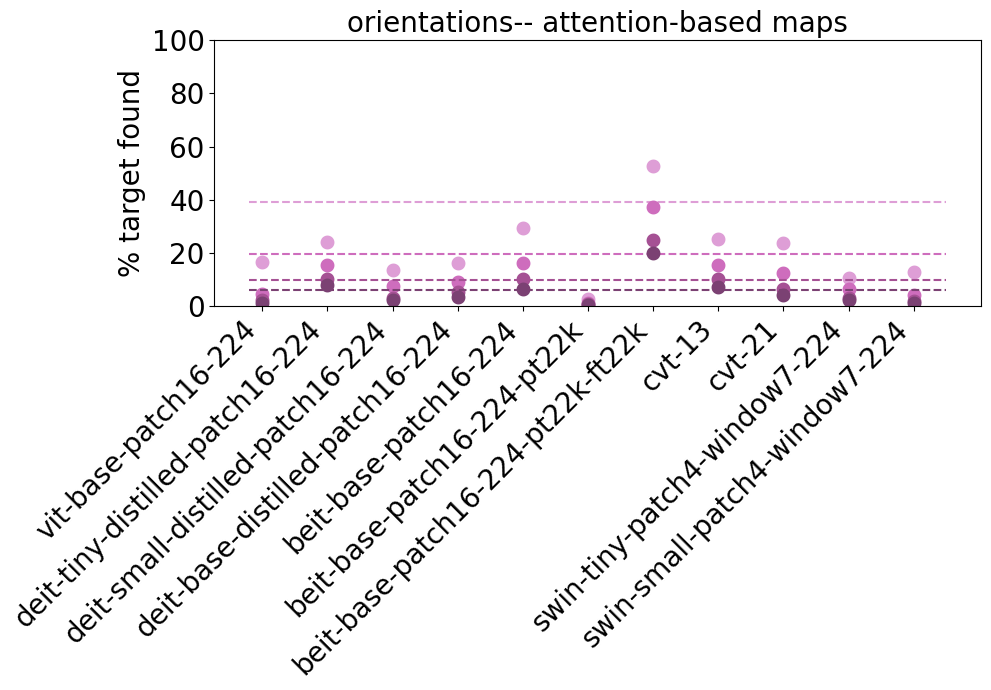}
	\end{subfigure}
	\begin{subfigure}{0.49\textwidth}
		\includegraphics[width=\linewidth]{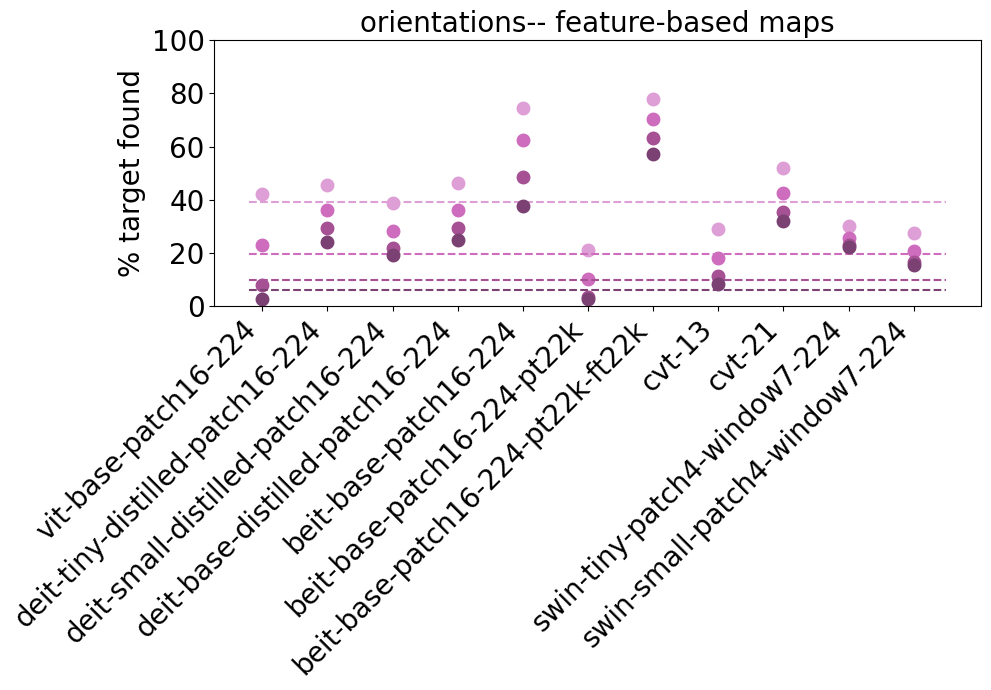}
	\end{subfigure}
	
	\begin{subfigure}{0.49\textwidth}
		\includegraphics[width=\linewidth]{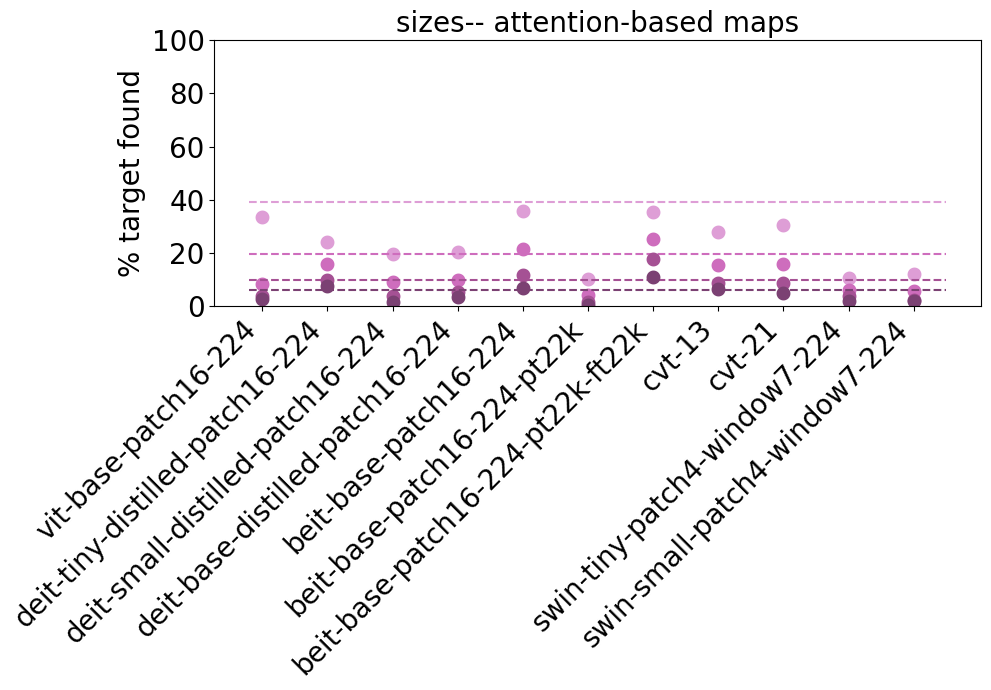}
	\end{subfigure}
	\begin{subfigure}{0.49\textwidth}
		\includegraphics[width=\linewidth]{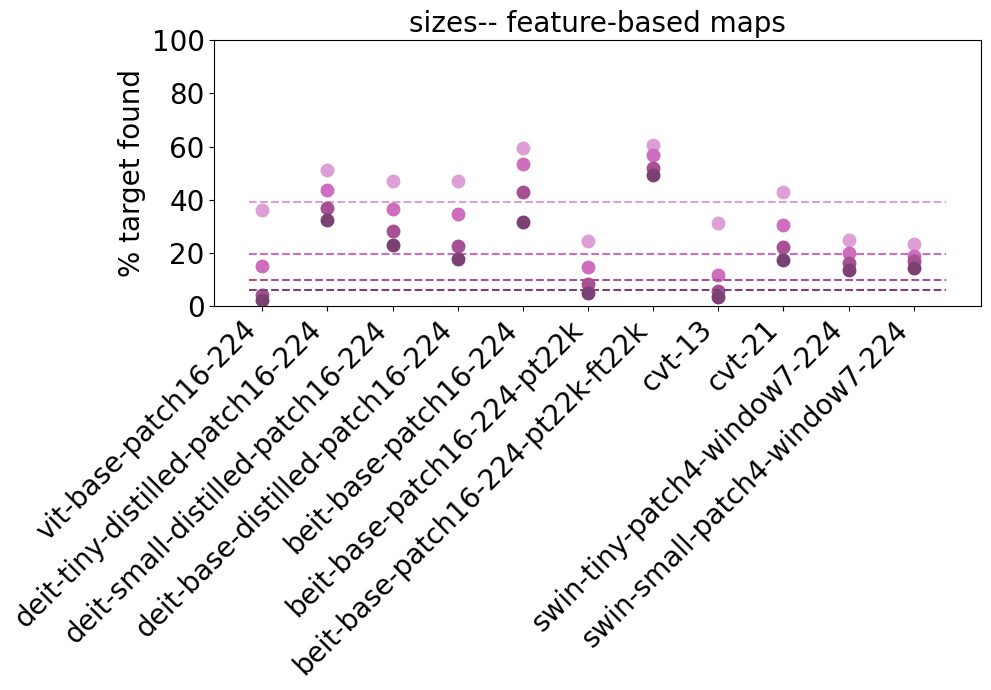}
	\end{subfigure}
	\caption{Target detection rate of the last block of vision transformers investigated in this work for 15, 25, 50, and 100 fixations on stimuli from the $\text{P}^3$ dataset. The chance level performance of the ViT model is plotted as dashed lines with matching colors for each fixation threshold. Similar to the observation of ViT, feature-based maps outperform attention-based maps and generally at rates higher than chance. Color targets are easier to detect for both map types. Interestingly, both Swin architectures struggle to detect size targets in both attention and feature-based maps, despite incorporating multiple scales in their model. }
	\label{fig:P3-last-layer}
\end{figure}

\begin{figure}
	\centering
	\begin{subfigure}{0.8\textwidth}
		\includegraphics[width=\linewidth]{P3_legend.png}
	\end{subfigure}  
	
	\begin{subfigure}{0.49\textwidth}
		\includegraphics[width=\linewidth]{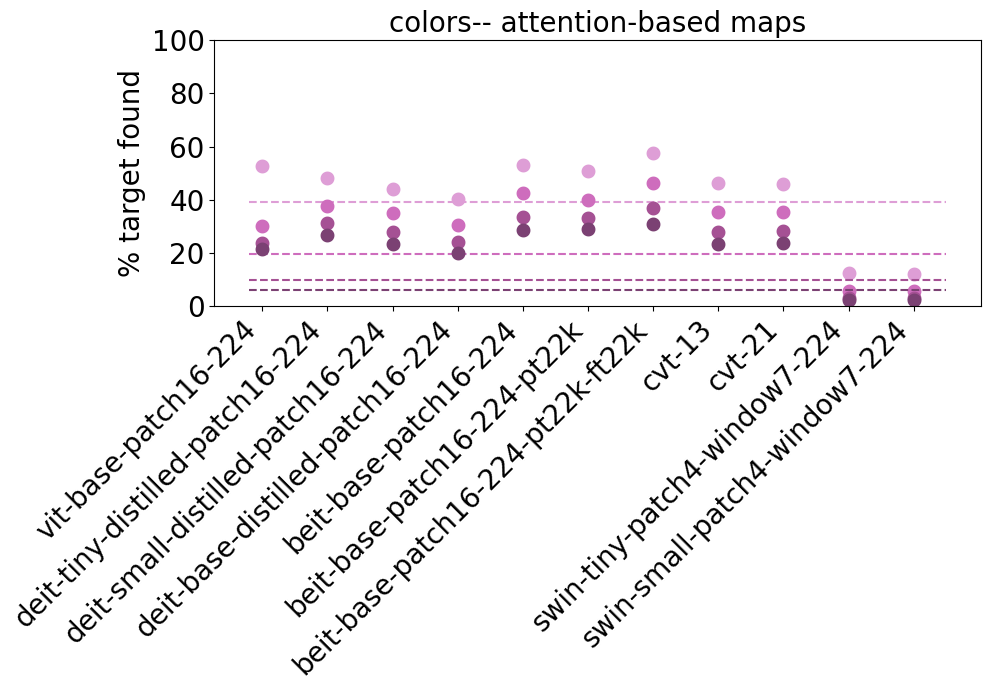}
	\end{subfigure}
	\begin{subfigure}{0.49\textwidth}
		\includegraphics[width=\linewidth]{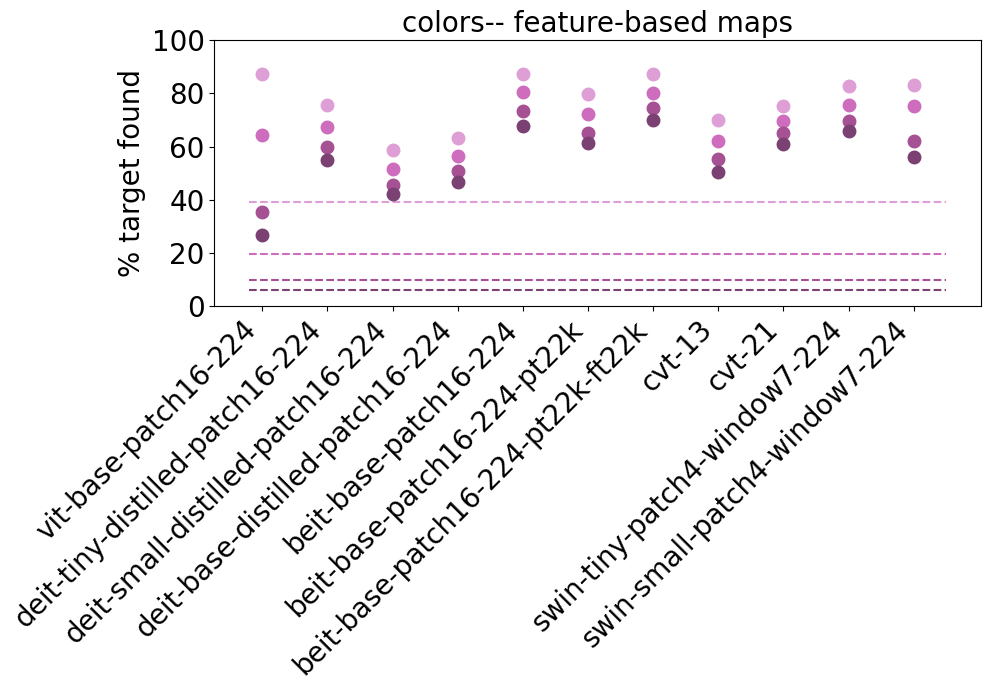}
	\end{subfigure}
	
	\begin{subfigure}{0.49\textwidth}
		\includegraphics[width=\linewidth]{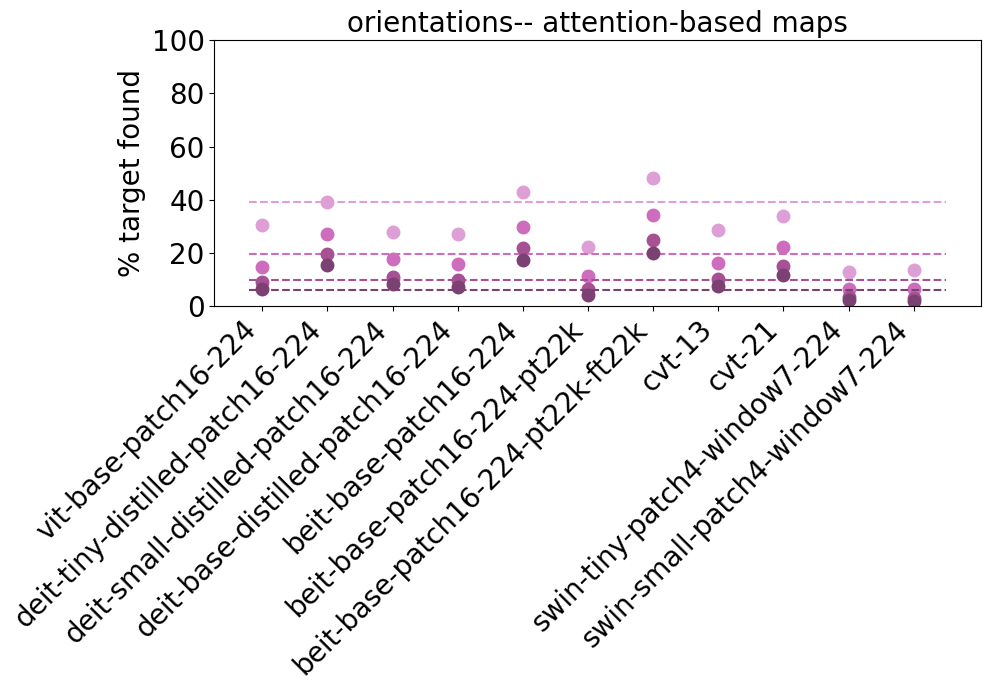}
	\end{subfigure}
	\begin{subfigure}{0.49\textwidth}
		\includegraphics[width=\linewidth]{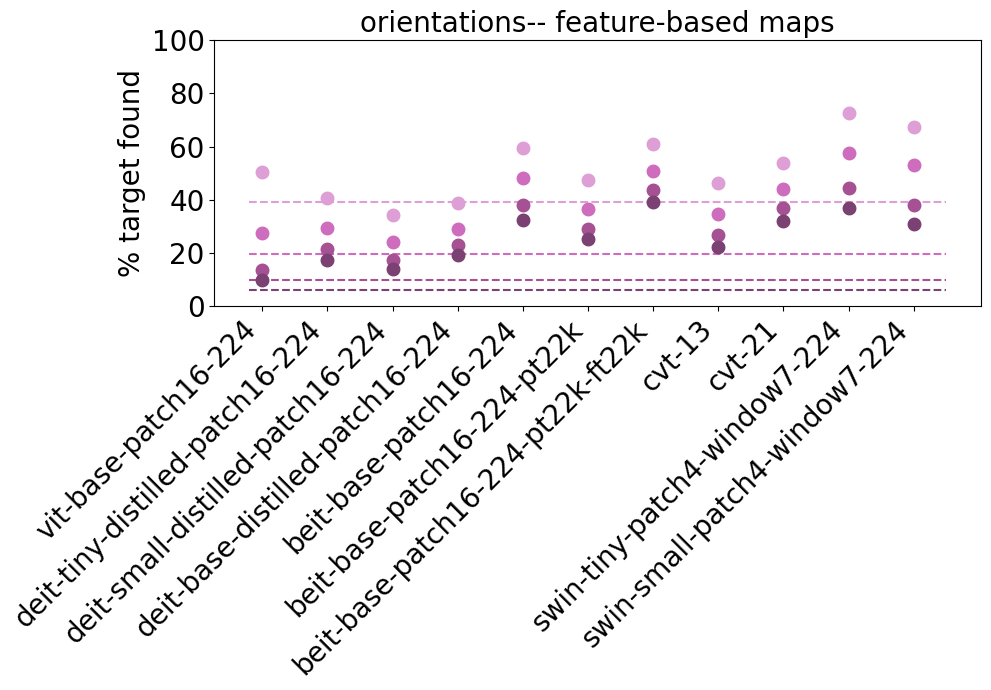}
	\end{subfigure}
	
	\begin{subfigure}{0.49\textwidth}
		\includegraphics[width=\linewidth]{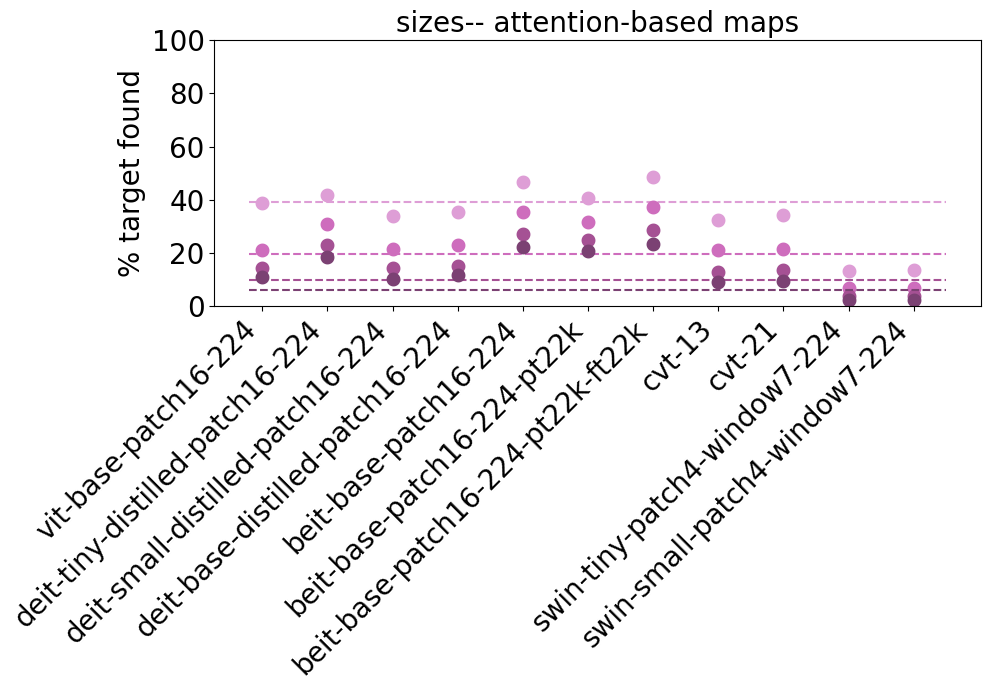}
	\end{subfigure}
	\begin{subfigure}{0.49\textwidth}
		\includegraphics[width=\linewidth]{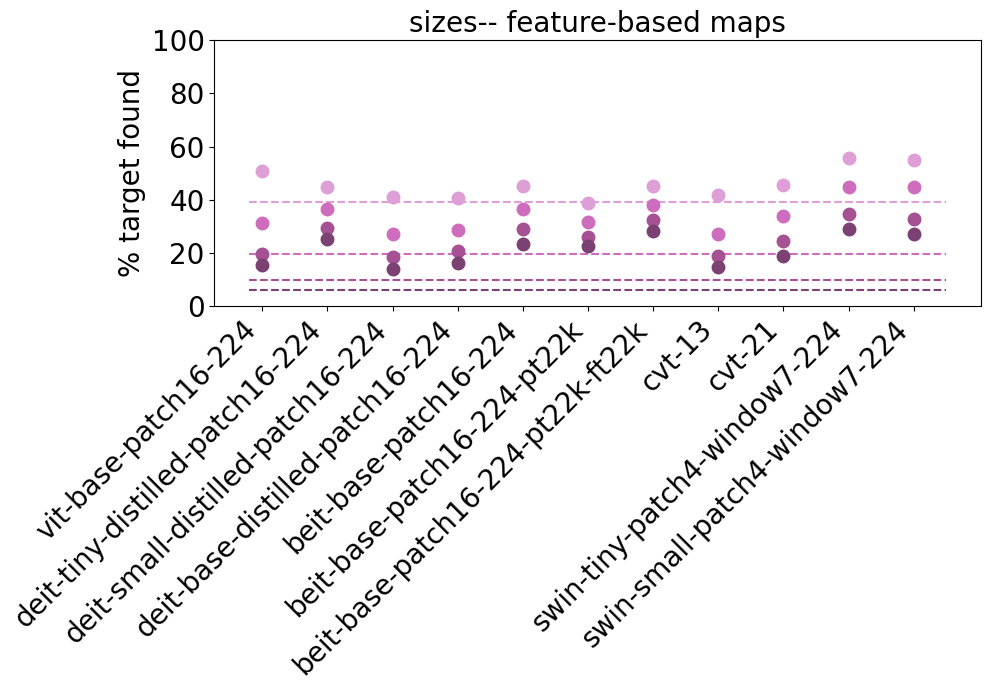}
	\end{subfigure}
	\caption{Average target detection rate over all blocks of vision transformers for 15, 25, 50, and 100 fixations on stimuli from the $\text{P}^3$ dataset. Compared to detection rates in Figure~\ref{fig:P3-last-layer}, mean detection rates are higher for all models in all conditions (color, size, and orientation), indicating superior performance of earlier transformer blocks compare to the final block in these models. In line with results in Figure~\ref{fig:P3-vit} and Figure~\ref{fig:P3-last-layer}, color targets are easier to detect and that generally, feature-based maps outperform attention-based maps in salient target detection.}
	\label{fig:P3-mean-layer}
\end{figure}

Similar detection rate patterns were observed in other vision transformer models. However, due to limited space, we refrain from reporting the same plots as in Figure~\ref{fig:P3-vit} for all the vision transformer models that we studied. These plots can be found in the Supplementary Material. Here, for each model, we report the mean target detection rate over all blocks and the detection rate for the last block of each model for both attention and feature-based saliency maps. These results are summarized in Figure~\ref{fig:P3-last-layer} and Figure~\ref{fig:P3-mean-layer} for the last and mean layer target detection rates respectively. Consistent with the observations from vit-base-patch16-224 in Figure~\ref{fig:P3-vit}, the feature-based saliency maps outperform attention-based ones in Figure~\ref{fig:P3-last-layer} and in general have higher detection rates than the chance levels stated earlier. The attention-based saliency maps, across most of the models, fail to perform better than chance. Generally, all models have higher detection rates for color targets, repeating similar results reported by Kotseruba~\etal~\cite{kotseruba2020saliency}. Interestingly, swin architectures that incorporate multiple token scales, perform poorly in detecting size targets with both feature and attention-based saliency maps. 

Results for mean target detection rates over all blocks in Figure~\ref{fig:P3-mean-layer} are comparable to those of last layer detection rates, except for a shift to higher rates. Specifically, all models are more competent at detecting color targets and that the feature-based saliency maps look more appropriate for singleton detection. In swin architectures, the mean detection rate of feature-based saliency maps are relatively higher for size targets than that of other models. This observation, together with the last layer detection rate of swin models for size targets suggest that incorporating multiple scales in vision transformers improves representing figures of various sizes but the effect fades higher in the hierarchy.

In summary, the attention maps in vision transformers were expected to reveal high salience for the target versus distractors. Nonetheless, comparing the detection rate of attention-based saliency maps in vision transformers at 100 fixations with those of traditional and deep saliency models reported by Kotseruba~\etal~\cite{kotseruba2020saliency} suggest that not only do the attention modules in vision transformers fail to highlight the target, but also come short of convolution-based deep saliency models with no attention modules. Although the feature-based saliency maps in vision transformers showed promising results in target detection rates relative to attention-based maps, in comparison with convolutional saliency models (See~\cite{kotseruba2020saliency}, their Figure 3), those performed relatively similar to convolution-based models. Together, these results suggest that contrary to the expectation, the proposed attention mechanisms in vision transformers are not advantageous versus convolutional computations in representing visual salience.

\subsubsection{The $\text{O}^3$ Dataset Results}
\begin{figure}
	\centering
	\begin{subfigure}{0.9\textwidth}
		\includegraphics[width=\linewidth]{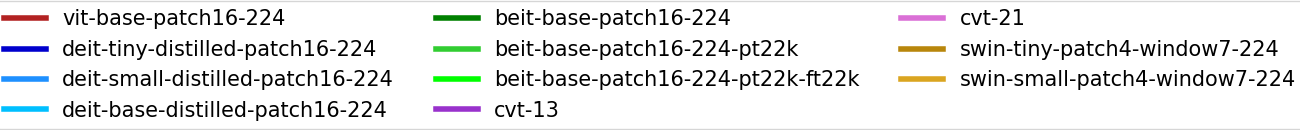}
	\end{subfigure}  
	
	\begin{subfigure}{0.49\textwidth}
		\includegraphics[width=\linewidth]{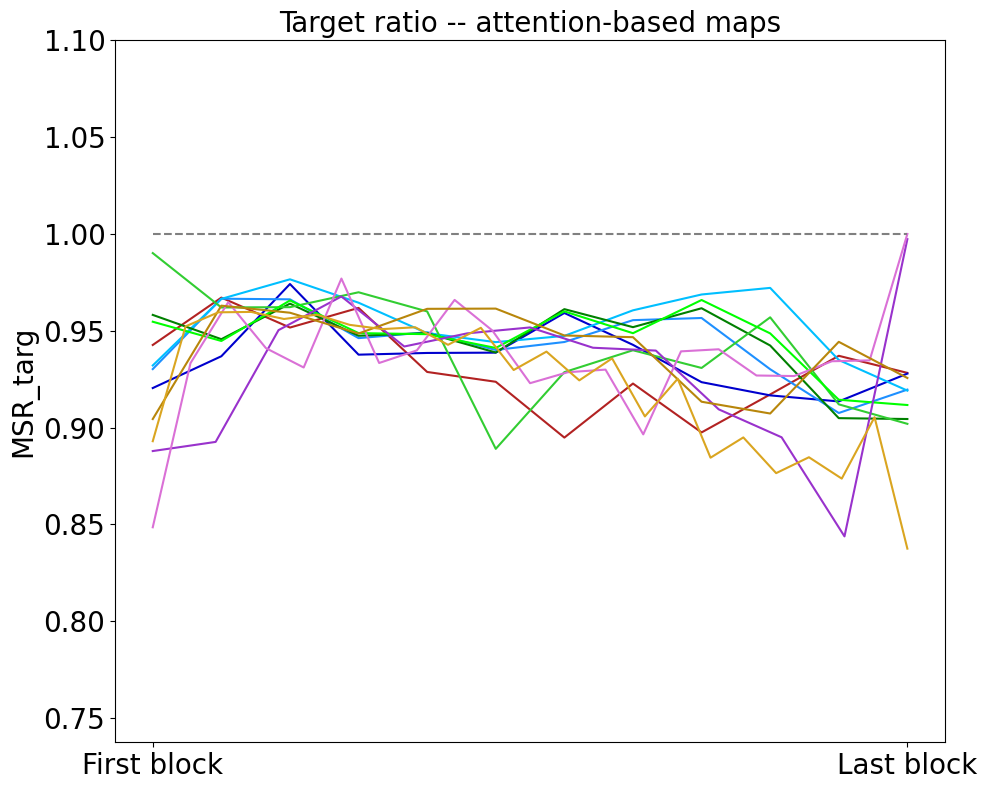}
	\end{subfigure}
	\begin{subfigure}{0.49\textwidth}
		\includegraphics[width=\linewidth]{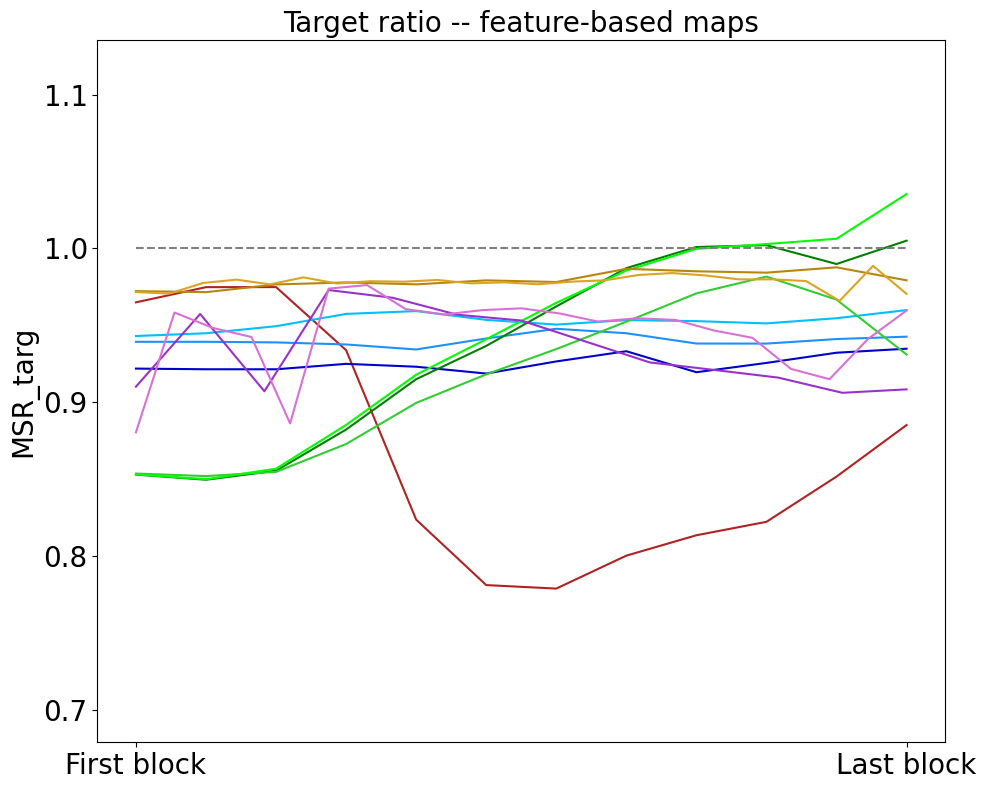}
	\end{subfigure}
	
	\begin{subfigure}{0.49\textwidth}
		\includegraphics[width=\linewidth]{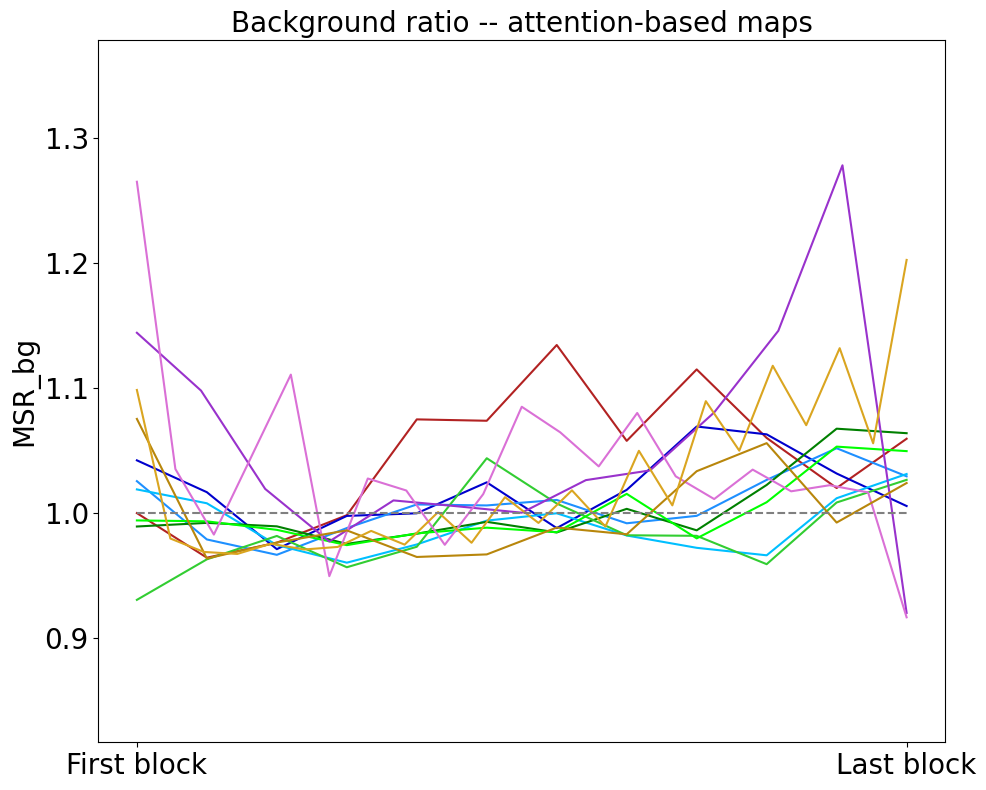}
	\end{subfigure}
	\begin{subfigure}{0.49\textwidth}
		\includegraphics[width=\linewidth]{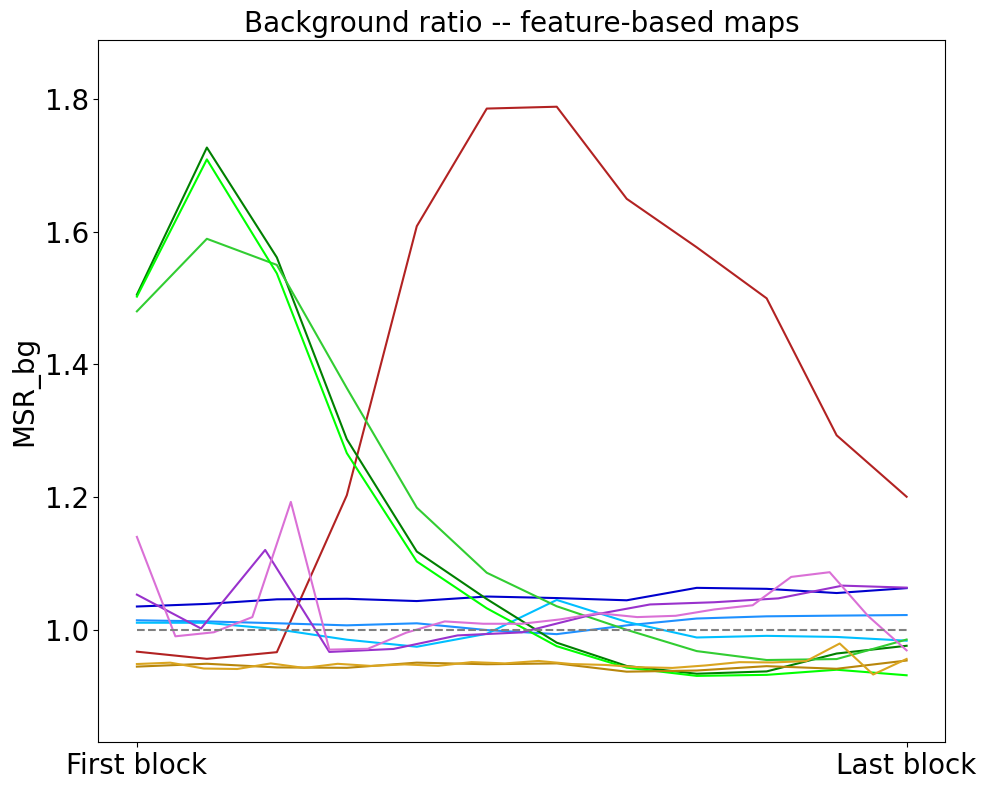}
	\end{subfigure}
	\caption{Target and background saliency ratios ($MSR_{targ}$ and $MSR_{bg}$) for all vision transformer architectures on natural images of the $\text{O}^3$ dataset. Even though these models were trained on natural images, they fail in assigning higher saliency to the target object versus the distractors or background (smaller than 1 $MSR_{targ}$ and larger than 1 $MSR_{bg}$ ratios) in both feature and attention-based maps. These results compared to the reported ratios in~\cite{kotseruba2020saliency} for traditional and deep convolution-based saliency models suggest that the proposed attention mechanisms do not enhance the performance of vision transformers when the goal is to detect the visually salient object in the stimulus.}
	\label{fig:O3-ratios}
\end{figure}

We measured the maximum saliency ratios $MSR_{targ}$ and $MSR_{bg}$ for feature and attention-based saliency maps of all blocks of vision transformers in Table~\ref{table:model_details}. These ratios are plotted in Figure~\ref{fig:O3-ratios}, demonstrating poor performance of all models in detecting the target in natural images of the $\text{O}^3$ dataset. We acknowledge that we expected improved performance of vision transformers on the $\text{O}^3$ dataset with natural images compared to the results on synthetic stimuli of the $\text{P}^3$ dataset. However, whereas $MSR_{targ}$ ratios larger than 1 are expected (higher salience of target versus distractors), in both feature and attention-based saliency maps, the ratios were distinctly below 1 across all blocks of all models, with the exception of later blocks of two beit architectures. Notable are the feature-based ratios of vit-base-patch16-224 with peaks in earlier blocks and a steep decrease in higher layers. In contrast, all three beit architectures show the opposite behavior and perform poorly in earlier blocks but correct the ratio in mid-higher stages of processing.

The $MSR_{bg}$ ratios illustrated in Figure~\ref{fig:O3-ratios} follow a similar theme as $MSR_{targ}$ ratios. Even though $MSR_{bg}$ ratios less than 1 suggest that the target is deemed more salient than the background, most of these models have $MSR_{bg}$ ratios larger than 1 in their hierarchy. Among all models, feature-based saliency of beit and swin architectures have the best overall performance. 

To summarize, for a bottom-up architecture that is claimed to implement attention mechanisms, we expected a boost in saliency detection compared to convolution-based models with no explicit attention modules. Our results on the $\text{O}^3$ dataset, however, point to the contrary, specifically in comparison with the best ratios reported in~\cite{kotseruba2020saliency} for $MSR_{targ}$ and $MSR_{bg}$ at 1.4 and 1.52 respectively. These results, together with the proposal of Liu~\etal~\cite{liu_convnet_2022} for a modernized convolution-based model with comparable performance to vision transformers, overshadow the claim of attention mechanisms in these models.

\section{Discussion}
Our goal in this work was to investigate if the self-attention modules in vision transformers have similar effects on visual processing as those reported in humans. Vision transformers have attracted much interest in the past few years partly due to out-performing CNNs in various visual tasks, and in part due to incorporating modules that were claimed to implement attention mechanisms. Despite some previous attempts~\cite{naseerintriguing,tuli2021convolutional,park_how_2022}, the role and effect of the attention modules in vision transformers have been largely unknown as, to the best of our knowledge, these studies did not investigate if the computations in self-attention modules would have similar effects on visual processing as those discovered with visual attention in humans. 

In this work, we studied two aspects of processing in vision transformers: the formulation of attention in self-attention modules, and the overall bottom-up architecture of these deep neural architectures. Our investigation of attention formulation in vision transformers suggested that these modules perform Gestalt-like similarity grouping in the form of horizontal relaxation labeling whereby interactions from multiple spatial positions determine the update in the representation of a token. Additionally, given previous evidence on the role of feedback in human visual attention~\cite{Peterson2014,desimone95,connor2004visual,baluch2011mechanisms,folk1992involuntary,bacon1994overriding,kim1999top,lamy2003does,yantis1999distinction}, we argued that if vision transformers implement attention mechanisms, those can only be in the form of bottom-up and stimulus-driven visual salience signals. 

Testing a family of vision transformers on a similarity grouping dataset suggested that the attention modules in these architectures perform similarity grouping and that the effect decays as hierarchical level increases in the hierarchy especially because more non-figure tokens are grouped with figures in the stimulus over multiple transformer encoder blocks. Most surprising, however, were our findings in the task of singleton detection as a canonical example of saliency detection. With both synthetic and natural stimuli, vision transformers demonstrated sub-optimal performance in comparison with traditional and deep convolution-based saliency models.

The $\text{P}^3\text{O}^3$ dataset was designed according to psychological and neuroscience findings on human visual attention. Kotseruba~\etal~\cite{kotseruba2020saliency} demonstrated a gap between human performance and traditional/CNN-based saliency models in singleton detection tasks. The fact that Kotseruba~\etal~\cite{kotseruba2020saliency} reported that training CNN-based saliency models on these stimuli did not improve their performance, hints on a more fundamental difference between the two systems. Several other works have provided evidence on the lack of human equivalence in deep neural networks~\cite{wloka2019flipped,xu2021examining,xu2021limits,xu2022tolerace,geirhos2018imagenettrained,ghodrati2014feedforward,webster_3dCNN,dodge,hu2019figure,kim2018not,horikawa2019characterization,zerroug2022a,Vaishnav,fel2022harmonizing,ricci2021same,baker2020local,ayzenberg2022perception,lonnqvist2021comparative,zhou2022exploringViT,feather2022metamers} on various aspects of visual processing. The claim of implementing attention mechanisms in vision transformers offered the possibility that these models might be more human-like. This impression was confirmed in the work of Tuli~\etal~\cite{tuli2021convolutional} who reported that vision transformers are more human-like than CNNs based on performance on the Stylized ImageNet dataset~\cite{geirhos2018imagenettrained}. Our work, however, adds to the former collection of studies and reveals a gap between human visual attention and the mechanisms implemented in vision transformers.

This work can be further extended in several directions. For example, even though Kotseruba~\etal~\cite{kotseruba2020saliency} found training CNN-based saliency models on the $\text{O}^3$ dataset did not improve their saliency detection performance, an interesting experiment is to fine-tune vision transformers on the $\text{O}^3$ dataset and evaluate the change or lack of change in their saliency detection performance. Additionally, incorporating vertical visual processes into the formulation in Equation~\ref{eq:attn} is another avenue to explore in the future.

To conclude, not only does our deliberate study of attention formulation and the underlying architecture of vision transformers suggest that these models perform perceptual grouping and do not implement attention mechanisms, but also our experimental evidence, especially from the $\text{P}^3\text{O}^3$ datasets confirms those observations. The mechanisms implemented in self-attention modules in vision transformers can be interpreted as \textit{lateral interactions} within a single layer. In some architectures, such as ViT, the entire input defines the neighborhood for these lateral interactions, in some others~\cite{focal_attn_2021} this neighborhood is limited to local regions of input. Although Liu~\etal~\cite{liu_convnet_2022} found similar performance in a modernized CNNs, the ubiquity of lateral interactions in the human and non-human primate visual cortex~\cite{shushruth2013different,stettler2002lateral} suggest the importance of these mechanisms in visual processing. These interactions in the biological visual system, among other effects, deal with noise and uncertainty in the input visual signal. Self-attention in vision transformers performs perceptual grouping, not attention. Additionally, considering Gestalt principles of grouping, vision transformers implement a narrow aspect of perceptual grouping, namely similarity, and other aspects such as symmetry and proximity seem problematic for these models. Unfortunately, the confusion caused by inappropriate use of terminology and technical term conflation has already been problematic. Therefore, we remain with the suggestion that even though vision transformers do not perform attention as claimed, they incorporate visual mechanisms in deep architectures that were previously absent in CNNs and provide new opportunities for further improvement of our computational vision models.

\section*{Acknowledgments}
Funding: This research was supported by several sources for which the authors are grateful: Air Force Office of Scientific Research [grant number FA9550-18-1-0054]; the Canada Research Chairs Program [grant number 950-231659]; and the Natural Sciences and Engineering Research Council of Canada [grant number RGPIN-2016-05352 and RGPIN-2022-04606].

\bibliographystyle{plain} 
\bibliography{transformer-grouping}

\end{document}